\documentclass[runningheads]{llncs}

\usepackage{eccv}

\usepackage{eccvabbrv}
\usepackage{appendix}
\usepackage{bbding}
\usepackage{booktabs}
\usepackage{graphicx}	
\usepackage{amsmath}	
\usepackage{amssymb}	
\usepackage{microtype}
\usepackage{xcolor, colortbl}
\usepackage{stfloats}
\usepackage{enumitem}
\usepackage{tabularx}
\usepackage{multirow}
\usepackage{wrapfig}
\usepackage{xspace}
\usepackage{url}
\usepackage{comment}
\usepackage{subcaption}
\usepackage{algorithm}
\usepackage{algorithmic}
\usepackage[accsupp]{axessibility} 
\usepackage{pifont}
\usepackage{utfsym}
\usepackage{makecell}
\usepackage{placeins}
\definecolor{Firebrick}{RGB}{178,34,34}
\definecolor{tabfirst}{rgb}{1.0, 0.86, 0.86} 
\definecolor{tabsecond}{rgb}{0.65, 0.92, 0.87}
\definecolor{tabthird}{rgb}{0.79, 0.92, 1.0}

\definecolor{GoogleBlue}{HTML}{1A73E8}
\usepackage[colorlinks=true,
            linkcolor=red,   
            citecolor=GoogleBlue,  
            urlcolor=GoogleBlue]{hyperref}

\usepackage{graphicx}
\usepackage{siunitx}
\usepackage{amsmath,mathtools}
\usepackage{breqn}
\usepackage{wrapfig}

\begin{document}

\setlength{\emergencystretch}{3em}
\title{The 3D Mirage: Probing and Taming \\ 3D Hallucinations} 

\author{
Hoang Nguyen\textsuperscript{*} \and
Xiaohao Xu\textsuperscript{*}${}^{\dagger}$ \and
Xiaonan Huang
}
\authorrunning{H. Nguyen et al.}

\institute{
University of Michigan, Ann Arbor, USA\\
\textsuperscript{*} Equal Contribution \quad ${}^{\dagger}$ Project Lead
}

\maketitle

\begin{abstract}
Monocular depth foundation models achieve remarkable generalization by learning large-scale semantic priors, but this creates a critical vulnerability: they hallucinate illusory 3D structures from planar/low-curvature but perceptually ambiguous inputs. We term this failure the \textbf{3D Mirage}. This paper introduces a novel end-to-end framework to \textbf{probe}, \textbf{score}, and \textbf{tame} this under-quantified safety risk \emph{in monocular depth under context variation}. To \textbf{probe}, we present \textbf{3D-Mirage}, the first benchmark to combine context variation and precise annotation for real-world illusions with real object exclusions, multi-surface support; \emph{purpose-built to stress-test monocular depth on real-world illusions}. To \textbf{score}, we propose a second-order magnitude-based evaluation with two metrics: the \textbf{Deviation Composite Score (DCS)} for high second-order 3D structure and the \textbf{Confusion Composite Score (CCS)} for contextual instability. To \textbf{tame} this failure, we introduce \textbf{Grounded Self-Distillation}, a parameter-efficient strategy on Depth-Anything-V2 baseline that surgically targets and resolves hallucination on illusion ROIs while preserving background knowledge, avoiding catastrophic forgetting. Our work provides an innovative pipeline for diagnosing and addressing this phenomenon, urging a necessary shift in the evaluation of MDE from pixel-wise accuracy to structural and contextual robustness.

\smallskip
\noindent\textbf{Code:}~\url{https://github.com/hdnndh/The-3D-Mirage-Probing-and-Taming-3D-Hallucinations}\\
\noindent\textbf{Dataset:}~\url{https://huggingface.co/datasets/3dmirage/3D-Mirage}
\end{abstract}


\section{Introduction}
\label{sec:intro}

\begin{wrapfigure}{R}{0.45\textwidth}%
    \centering
\includegraphics[width=0.45\textwidth]{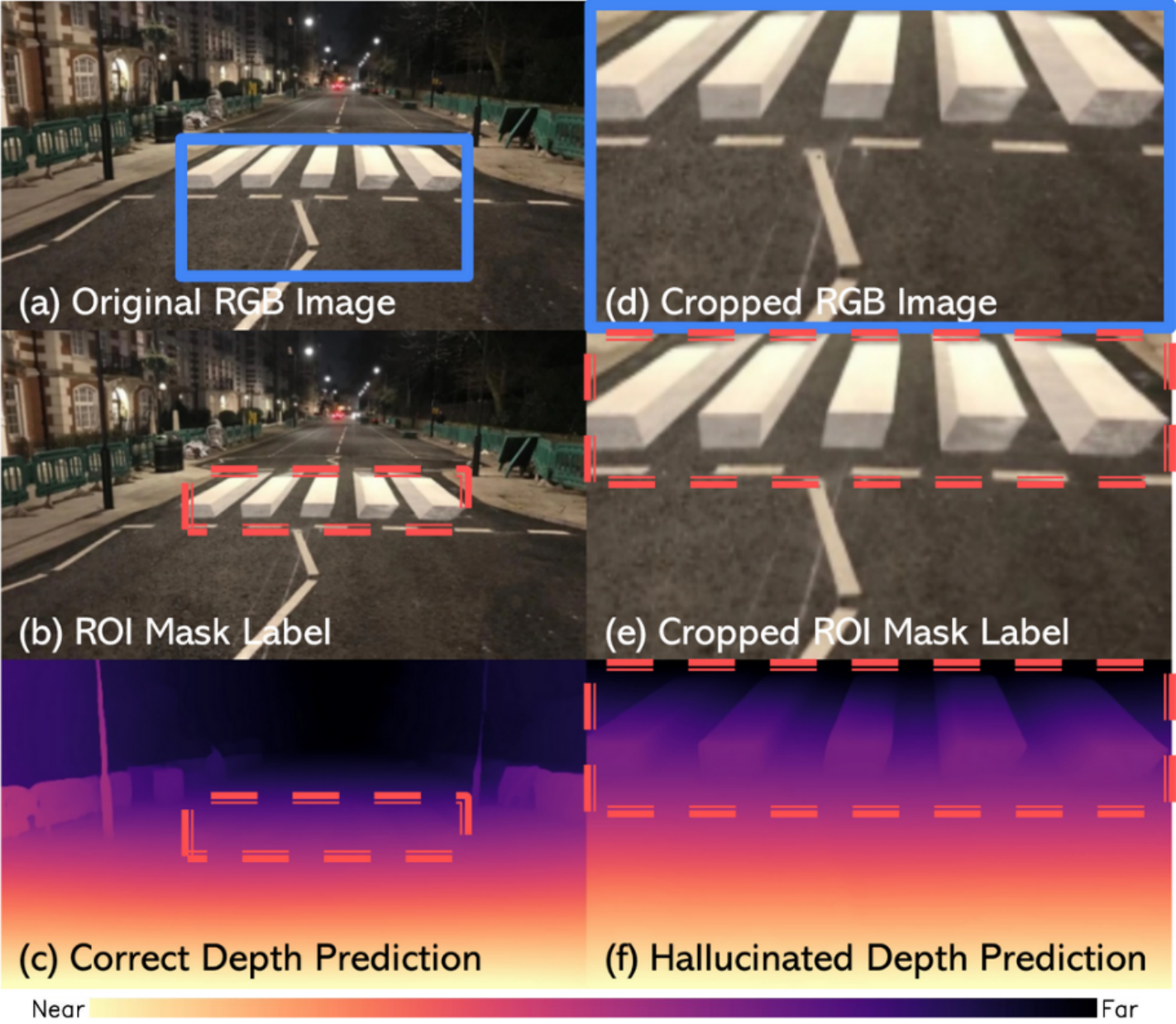}\vspace{-2mm}
\caption{
\textbf{The 3D Mirage: Hallucinations induced by illusory texture patterns.} (\textbf{a}) A driving scene featuring a deceptive non-existent 3D structure. (\textbf{b}) With full global context, the depth foundation model~\cite{dav2} correctly identifies the road as planar(\textbf{c}). (\textbf{d}-\textbf{f}) However, when the view is restricted to the local region, the model fails to disambiguate the texture from geometry. It hallucinates significant non-existent 3D obstacles (\textbf{f}) from the phantom pattern, illustrating a critical vulnerability in reliable 3D perception for autonomous driving scenarios.
}
    \label{fig:teaser}
\vspace{-8mm}
\end{wrapfigure}

Enabling reliable perception and reconstruction of 3D scenes is critical for safe and robust visual intelligence~\cite{Li_2023_ICCV,Li_2024_CVPR,li-etal-2023-towards-noise,xu2025scalable,li2022online,li2024r,qiu2025robust,qiu2025image} and autonomous driving experience~\cite{li2024optimizing}. Driven by this necessity, Monocular Depth Estimation (MDE) has transitioned from a challenging academic problem to a core perception component in real-world systems. This rapid adoption is fueled by powerful foundation models such as Depth-Anything V2~\cite{dav2}, Zoe-Depth~\cite{zoe}, and MiDaS/DPT~\cite{dpt, ranftl2022tpami}, which are trained on massive, diverse datasets. However, their remarkable zero-shot generalization obscures a critical and unexamined vulnerability: an over-reliance on large-scale statistical priors, trading geometric fidelity for semantic consistency, making them susceptible to perceptual ambiguity and adversarial attack.

In this work, we identify and analyze a critical failure mode we term the \textbf{3D Mirage}. We find that SOTA depth foundation models fail in two common, safety-critical scenarios: 1) when presented with perceptually ambiguous 2D patterns, such as 3D street art, and 2) when operating under a restricted field-of-view (FOV) that removes broad contextual cues. Fig.~\ref{fig:teaser} shows this: the same flat road section reads as planar in the full scene, yet yields a significant non-existent 3D structure once the view is restricted to it. The trigger of this phenomenon is the illusory texture on the low-curvature carrier surface, causing hallucination of 3D structure. Furthermore, a faithful predictor should report the same flat geometry however the context varies, and the change in context exposes a core dependency of hallucination. This is a failure of contextual grounding: the model's depth prediction is not anchored in local geometric reality, but is instead a fragile artifact of priors shaped by large-scale training.

This failure is not an isolated anecdote. We demonstrate that this vulnerability is systemic across the current generation of leading models. As shown in Fig.~\ref{fig:sota_hallucination}, we subjected a wide range of architectures, from transformer-based (Depth-Anything V2~\cite{dav2}) and diffusion-based (Marigold~\cite{Ke_2024_CVPR}) to generative (DepthFM~\cite{Gui_2025_AAAI}) and commercially-developed (Depth Pro~\cite{Bochkovskii_2025_ICLR}), to these 3D mirage inputs. All models exhibited similar failures, unstably predicting spurious 3D structures from low-curvature surfaces.

This collective failure exposes a critical gap in \textit{how} we evaluate these models. Standard metrics like Mean Absolute Error (MAE) and Root Mean Squared Error (RMSE)~\cite{rmsemae2015} are \textit{perceptually-blind} to these structural failures. By averaging pixel-wise errors, they cannot differentiate between a slight, uniform mis-calibration and a massive, hallucinated obstacle. We posit that MDE evaluation must evolve to assess \textbf{\textit{structural integrity}} and \textbf{\textit{contextual stability}}, which are far more critical for real-world deployment than pure pixel accuracy.

To address this, our work provides the first end-to-end framework to systematically \textbf{probe}, \textbf{score}, and \textbf{tame} 3D hallucinations. Our contributions are threefold:
\begin{itemize}
    \item We \textbf{probe} this vulnerability by introducing \textbf{3D-Mirage}, a benchmark purpose-built for paired full vs. context-varied evaluation of monocular depth on planar/low-curvature illusion carriers, with manually annotated ROI polygons supporting \textbf{\textit{multiple illusions}} per scene, \textbf{\textit{real object exclusions}}, and \textbf{\textit{illusion spanning surfaces}}.
    
    \item We \textbf{score} these failures by proposing a novel \textbf{Second-order magnitude-based evaluation framework}, introducing two metrics: the \textbf{Deviation Composite Score (DCS)} to measure spurious second-order structure (hallucination intensity) and the \textbf{Confusion Composite Score (CCS)} to measure contextual instability (\textit{i.e.}, the mirage effect).
    
    \item We \textbf{tame} these hallucinations with a novel \textbf{Grounded Self-Distillation} (GSD) strategy. By applying low-parameter adapters to the model's encoder, we use our benchmark to suppress spurious curvature on illusion ROIs while using the frozen teacher model to enforce alignment on stable background and border regions. This efficiently grounds the model, mitigating hallucinations without catastrophic forgetting of its core pre-trained knowledge.
\end{itemize}

Ultimately, our contributions provide the essential tools: a targeted benchmark, perceptually-aware metrics, and an efficient mitigation strategy to advance MDE from simple geometric accuracy to the structural and contextual robustness demanded by safety-critical applications.


\section{Related Works}
\label{sec:rel}

\subsection{2D and 3D Visual Hallucination}
\begin{wrapfigure}{r}{0.5\textwidth}%
\vspace{-8mm}
    \centering
    \includegraphics[width=0.5\textwidth]{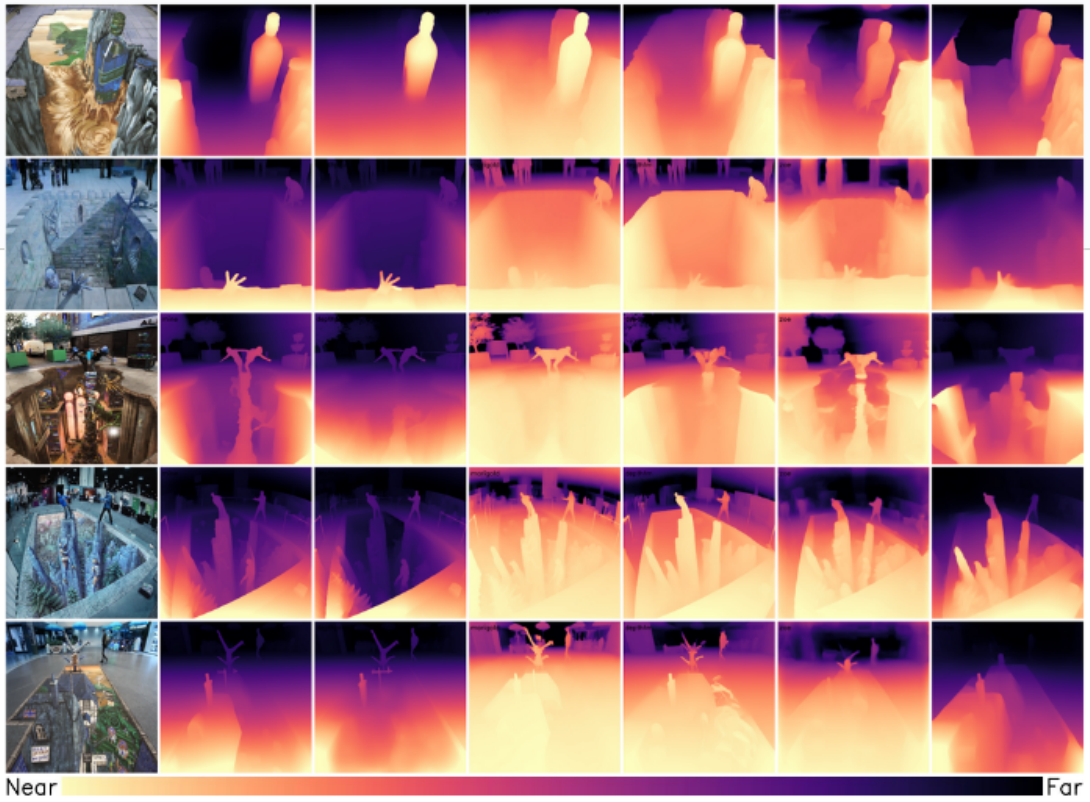}\vspace{-2mm}
    \caption{
    \textbf{Hallucinations across SOTA monocular depth models on images.}
    Given an optical-illusion region or a view with restricted context, all tested monocular depth foundation models
    (DAv2~\cite{dav2}, Depth Pro~\cite{Bochkovskii_2025_ICLR}), 
    Marigold~\cite{Ke_2024_CVPR},
    DepthFM~\cite{Gui_2025_AAAI}, ZoeDepth~\cite{zoe}, MiDaS~\cite{ranftl2022tpami},
    predict \emph{spurious depth variation/3D structure}. 
    }
    \label{fig:sota_hallucination}
    \vspace{-6mm}
\end{wrapfigure}
\textcolor{black}{Visual hallucination, predicting content not present in the input, is a known failure in 2D. This includes classifying nonsense images~\cite{nguyen2015deep}, detecting objects in empty locations~\cite{kayhan2021hallucination, moosavi2017CVPR}, or segmenting non-existent structures~\cite{lee2024volcano, chen2023iccv}. Such failures, perilous in safety-critical domains~\cite{leng2024vcd, zheng2024cvpr}, are linked to over-parameterization and models overly relying on context over evidence~\cite{kim2025segmentation, torralba2011cvpr, singh2020cvpr, chen2023iccv}.
In 3D, this problem is less studied but more complex. We define 3D hallucination as predicting depth variations on geometrically flat or low-curvature surfaces~\cite{muhovic2023hallucinating}. This is exacerbated by the ill-posed nature of MDE: the 3D-to-2D projection discards depth~\cite{saxena2005neurips}, forcing networks to use learned priors to resolve ambiguity. This can yield multiple valid reconstructions~\cite{chawla2021error, bian2019neurips} or overfitting to dataset- or camera-specific biases like texture cues~\cite{chen2016neurips}. Consequently, most MDE literature has focused on geometric accuracy rather than characterizing these structural failures.}

\subsection{MDE Models and Benchmarks}

Monocular Depth Estimation (MDE) seeks to recover 3D structure from a single RGB image. Early methods evolved from supervised~\cite{eigen2014nips} to self-supervised using geometric constraints~\cite{zhou2017cvpr, godard2017cvpr, godard2019iccv}. The field has recently shifted toward large foundation models like Depth Anything (DAv2)~\cite{dav1, dav2}, ZoeDepth~\cite{zoe}, MiDaS/DPT~\cite{ranftl2022tpami}, and Marigold~\cite{Ke_2024_CVPR}, Depth Pro~\cite{Bochkovskii_2025_ICLR}, DepthFM~\cite{Gui_2025_AAAI}. Trained on broad data, these models achieve remarkable zero-shot generalization but rely heavily on statistical priors. This reliance enables them to fill in depth in ambiguous or deceptive regions~\cite{zheng2024cvpr, wong2020neurips, xu2025towards}, trading geometric fidelity for semantic robustness.

\begin{wrapfigure}{r}{0.45\textwidth}%
\vspace{-8mm}
    \centering
    \includegraphics[width=\linewidth]{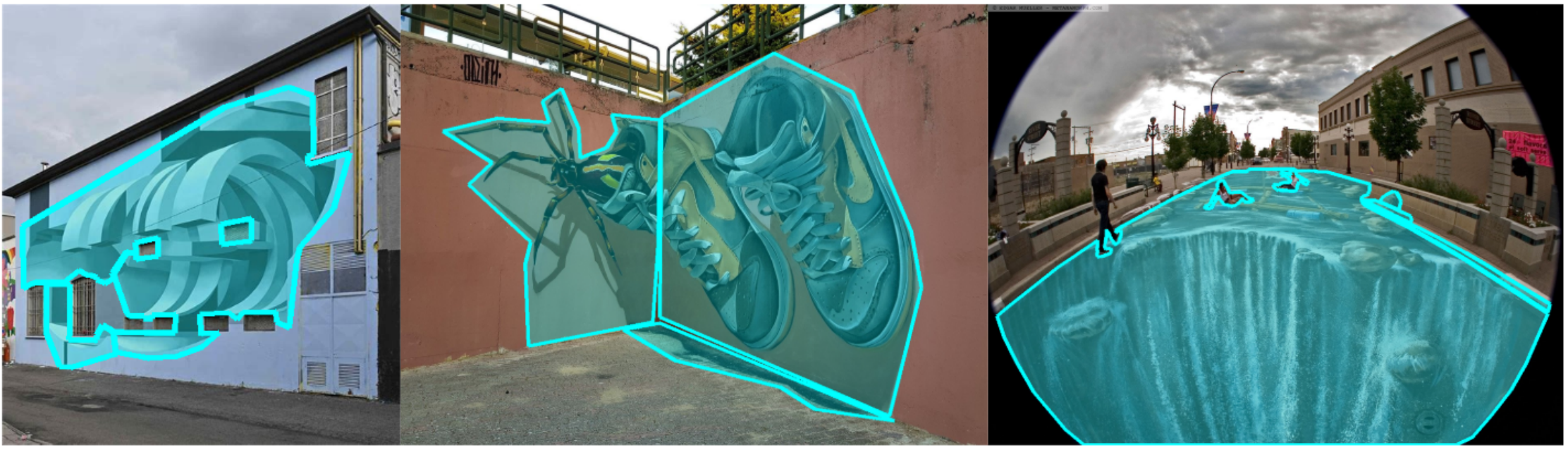}\vspace{-2mm}
    \caption{\textcolor{black}{\textbf{Visualization of ROI annotation}. We annotate the hallucination regions with individual polygon per surface \textit{\textbf{(2,3)}} and nested rule to exclude real objects \textit{\textbf{(1,3)}}. This works well with our plane-mixture and gating process in training.}}
    \label{fig:anno_vis}
    \vspace{-8mm}
\end{wrapfigure}

However, existing MDE benchmarks are insufficient for probing this failure mode. Mainstream datasets (KITTI ~\cite{kitti}~\cite{menze2015object}~\cite{geiger2012cvpr}, NYUv2~\cite{nyu}, ScanNet~\cite{scannet}, Depth-in-the-Wild~\cite{chen2016neurips}) emphasize \emph{geometrically-consistent} scenes, lacking the "perceptual traps" to trigger 3D hallucinations. Adversarial datasets are also limited: TartanAir-Adv~\cite{tartanair} uses synthetic motion. Other illusion datasets have different emphases: Booster~\cite{booster} focuses on specular/transparent surfaces, while MonoTrap~\cite{stereo_anywhere_cvpr2025} contains a limited indoor stereo setting (26 scenes). These illusion datasets lack systematic FOV variation.
Yao et al.~\cite{3dvi} recently introduced the 3D Visual Illusion Depth Estimation dataset and a VLM-driven framework that fuses binocular disparity with monocular depth. While this fusion improves robustness against visual illusions, it incurs substantial computational overhead, primarily from VLM. Furthermore, their results indicate that standard stereo baselines already perform exceptionally well on texture-rich, non-specular/transparent illusions (e.g., inpainting, pictures, replays, and holography) because of cross-view texture consistency. Ultimately, their primary advantage lies in using learned monocular priors to resolve mirror geometries. Their dataset focuses on 2D planar surface and evaluates on small-scale table-top and indoor wall scenes~\cite{3dvi}. In contrast, our benchmark consists of 468 indoor and outdoor texture illusions, both small and large scale scenes. We also specifically target a complementary failure mode in \emph{monocular} foundation models: \emph{context variation} (paired full-context and context-altered views) that induces unstable 3D hallucinations on low-curvature regions.
To our knowledge, \textbf{3D-Mirage} is the first benchmark centered on real-world low-curvature-carrier optical-illusion scenes that introduces controlled context variation (paired full-context and context-altered views) together with precise annotation for complex illusion cases (Fig.~\ref{fig:anno_vis}) to systematically evaluate hallucination and contextual stability in monocular depth.

\subsection{Probing and Mitigating 3D Hallucination}
\textcolor{black}{Early probes of MDE hallucination used textured transparent surfaces~\cite{costanzino2023learning} or scored failures from a semantic angle~\cite{lee2024volcano, lovenia2024nope, zheng2024cvpr}. However, these methods rarely localize the hallucination or quantify it systematically. Existing defenses are often model-specific and generalize poorly~\cite{huang2024cvpr, leng2024vcd, guizilini2020CVPR, heo2018ECCV}. To date, 3D hallucination remains difficult to label and formally quantify~\cite{muhovic2023hallucinating}, limiting systematic study.
While hallucinated 3D content can be viewed as a form of 3D anomaly or out-of-distribution 3D content, current approaches focus mainly on geometric 3D anomaly detection. Semantic 3D anomalies remain underexplored, and this work aims to address such semantic anomalies.
Given the scale of modern foundation models, full fine-tuning to correct such failures is prohibitive and risks catastrophic forgetting. Parameter-Efficient Fine-Tuning \cite{houlsby2019parameter} (PEFT) methods like Low-Rank Adaptation (LoRA)~\cite{hu2021lora} offer an alternative. LoRA freezes the model and injects small, trainable low-rank matrices, allowing efficient adaptation. We are the first to explore PEFT to tame 3D hallucinations with Depth-Anything-V2 as our baseline. We hypothesize that using a targeted benchmark, we can employ LoRA to ground a depth model, teaching it to ignore illusory 2D cues while preserving its pre-trained knowledge.}

\section{The 3D-Mirage Benchmark}
\label{sec:benchmark}

\textcolor{black}{To systematically probe the `3D Mirage' vulnerability, we introduce \textbf{3D-Mirage}, a benchmark purpose-built to elicit and measure 3D hallucinations in monocular depth models under \emph{illusory} and \emph{context-varied} conditions. The benchmark is designed not to test average-case accuracy, but to specifically target the failure modes where learned priors override geometric reality.}

\subsection{Dataset: Curation and Properties}
\textcolor{black}{The creation of 3D-Mirage involved a three-stage pipeline:}

\noindent\textbf{Data Collection.} \textcolor{black}{We first collected 468 real-world RGB images featuring \emph{painting} and \emph{street-art} 3D illusions across varied scenes. These include chalk anamorphoses, forced-perspective murals, and large-format advertisements that create a strong perceptual suggestion of 3D geometry on a 2D plane. Approximately 80\% of scenes are outdoor, 40\% lie on pedestrian walkways, 8\% are billboard/advertisement-like cases, 16\% span multiple support surfaces, and about 40\% contain real objects or people near or on top of the illusion.}

\noindent\textbf{ROI Annotation.} \textcolor{black}{After filtering, we manually annotated precise polygonal Region of Interest (ROI) masks for each illusion. If real objects are inside illusion, nested ROIs are used to mark and exclude them. These masks delineate regions that are low-curvature support surfaces yet \emph{suggest 3D structure in appearance}. This mask is the key component for our geometry-based evaluation.}

\noindent\textbf{Context-Variation Augmentation.} \textcolor{black}{To emulate the limited FOV and partial occlusions common in autonomous driving, we generated four random crops for each sample. These crops are centered on the ROI, with the illusion ROI covering at least $40\%$ of each crop's area, ensuring the illusion is present but the surrounding scene context is partially or fully absent.}

\begin{wrapfigure}{R}{0.45\textwidth}
\vspace{-8mm}
    \centering
    \includegraphics[width=\linewidth]{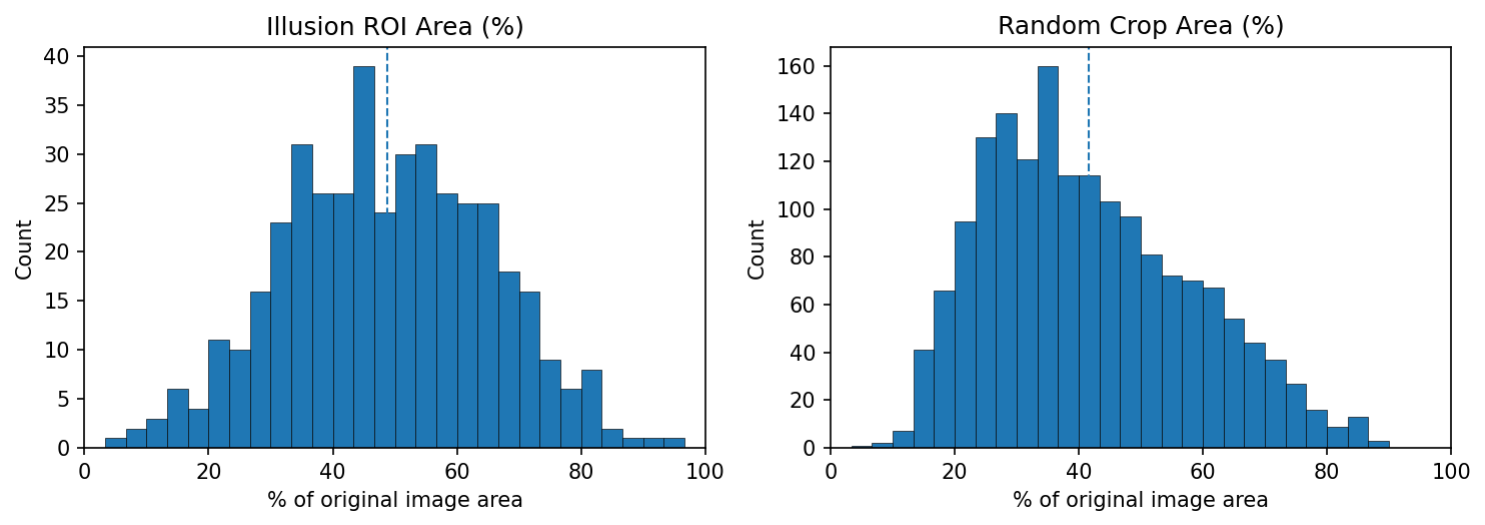}\vspace{-2mm}
    \caption{\textcolor{black}{\textbf{Statistics of illusion regions in the 3D-Mirage dataset}. Area distributions for illusion regions (left) and their corresponding random crops (right), as a percentage of the original image area. The dotted vertical line denotes the average value.}}
    \label{fig:data_stats}
    \vspace{-8mm}
\end{wrapfigure}

\noindent\textbf{Statistics.} \textcolor{black}{Each benchmark instance consists of a full-context image, its illusion ROI mask(s), and one of its four associated context-varied crops. The final benchmark contains \textbf{1,872} full-crop instances, all annotated and verified by human annotators. The dataset is designed to provide a challenging test of model robustness. As shown in Fig.~\ref{fig:data_stats}, the illusion ROIs are a significant part of the image, covering an average of 49\% of the total area. The context-varied crops are tighter, covering an average of 41\% of the original image.}

\subsection{Evaluation: Quantifying Hallucinations}
\label{sec:metrics}
\textcolor{black}{A core component of our benchmark is an evaluation framework that complements standard metrics, especially when \textbf{\textit{GT is not available or hard to collect at scale}}, to quantify hallucination and confusion. DCS/CCS are not intended to replace ground-truth depth metrics; they are reference-free structural probes for the annotated low-curvature carrier regions and are used together with standard depth and preservation evaluations in Sec.~\ref{sec:experiments}.}

\noindent\textbf{Shortcomings of Standard Metrics.}
\textcolor{black}{Standard metrics (MAE, RMSE, REL) dilute ROI-specific failures and evaluate views independently. \emph{Even when restricted to the ROI, they require GT depth and do not measure full--crop consistency for the same physical region.} Scale-invariant (SI-log) and AbsRel objectives share this gap; they score pixel-wise agreement with ground-truth depth, not the \emph{second-order curvature} that defines a 3D mirage, so a hallucinated bulge that preserves ordinal depth can satisfy them; DCS and CCS instead quantify ROI curvature directly and without GT.}

\noindent\textbf{Dual-View Projection Space.}
\textcolor{black}{Let $f_\theta$ be an MDE model. For each benchmark instance $i$ with full image $x_{\text{full}}$, crop $x_{\text{crop}}$ defined by crop box $b_i$, and ROI polygons, we compute $D_{\text{full}}=f_\theta(x_{\text{full}})$ and $D_{\text{crop}}=f_\theta(x_{\text{crop}})$. Using $b_i$, we crop and resize the full-view depth prediction to the crop frame, obtaining $\tilde D_{\text{full}\rightarrow\text{crop}}$, and rasterize the union of ROI polygons into a crop-frame mask $M_i$. For context variation, the full-image branch $D_{\text{full}}$ is itself a \emph{context-expansion baseline} of the crop view. An image-edited, non-illusory counterpart at matched scene context and crop support scores only $\mathrm{DCS}\,{\approx}\,54$ and $\mathrm{CCS}\,{\approx}\,1.4\times10^{-4}$, with added Gaussian noise lifting DCS to at most ${\approx}\,65$, whereas the illusion reaches $\mathrm{DCS}\,{\approx}\,861$ and $\mathrm{CCS}\,{\approx}\,7.4\times10^{-4}$ (a ${\sim}16\times$ DCS gap). DCS and CCS therefore stay insensitive to alignment artifacts, crop geometry, and mild pixel-level noise, and the full control study is in the supplementary material.}

\noindent\textbf{ROI-normalized Second-order magnitude-based responses.}
\textcolor{black}{Let $\mathrm{QN}(\cdot\,;M)$ denote ROI-conditional 1--99\% quantile normalization (with quantiles computed on ROI pixels) and let $L(\cdot)$ be the second-order magnitude operator. We define
\vspace{-1mm}
\[
R^{\text{full}}_i=L(\mathrm{QN}(\tilde D_{\text{full}\rightarrow\text{crop}};M_i)), 
\qquad
R^{\text{crop}}_i=L(\mathrm{QN}(D_{\text{crop}};M_i)).
\]
For $v\in\{\text{full},\text{crop}\}$, $R_i^{v}$ denotes the ROI-normalized second-order response map for branch $v$. From $R_i^{v}$, we compute two ROI scalars within $M_i$: \textbf{top10sum} $t_{v,i}$, sum of the largest 10\% response values; and \textbf{mean10} $\mu_{v,i}$, a 10\%-trimmed high-response mean computed over the largest 90\% response values.}

\noindent\textbf{Deviation Composite Score (DCS).}
\textcolor{black}{Let $\mathbf{t}_i=\big[t_{\text{full},i},\,t_{\text{crop},i}\big]^\top\in\mathbb{R}^2$ and $\bar{\mathbf{t}}=\frac{1}{N}\sum_i \mathbf{t}_i$. We define
\vspace{-1mm}
\begin{equation}
d_{\text{cluster}}=\|\bar{\mathbf{t}}\|_2,\qquad
d_{\text{avg}}=\frac{1}{N}\sum_i \|\mathbf{t}_i\|_2,\qquad
\mathrm{DCS}=d_{\text{cluster}}+d_{\text{avg}}.
\end{equation}}
\noindent\textbf{Confusion Composite Score (CCS).}
\textcolor{black}{Let $\boldsymbol{\mu}_i=\big[\mu_{\text{full},i},\,\mu_{\text{crop},i}\big]^\top$ and $\bar{\boldsymbol{\mu}}=\frac{1}{N}\sum_i \boldsymbol{\mu}_i$. Let $\mathbf{u}=\frac{1}{\sqrt{2}}[1,-1]^\top$ be the off-diagonal unit direction. We define
\begin{equation}
D_{\text{cluster}}=\big|\mathbf{u}^\top \bar{\boldsymbol{\mu}}\big|,\qquad
D_{\text{avg}}=\frac{1}{N}\sum_i \big|\mathbf{u}^\top \boldsymbol{\mu}_i\big|,\qquad
\mathrm{CCS}=D_{\text{cluster}}+D_{\text{avg}}.
\end{equation}}

A low DCS indicates little spurious second-order structure inside the annotated low-curvature carrier, while a low CCS indicates that the same physical region remains stable between full and context-restricted views.

\begin{figure*}[t]
  \centering
  \includegraphics[width=\textwidth]{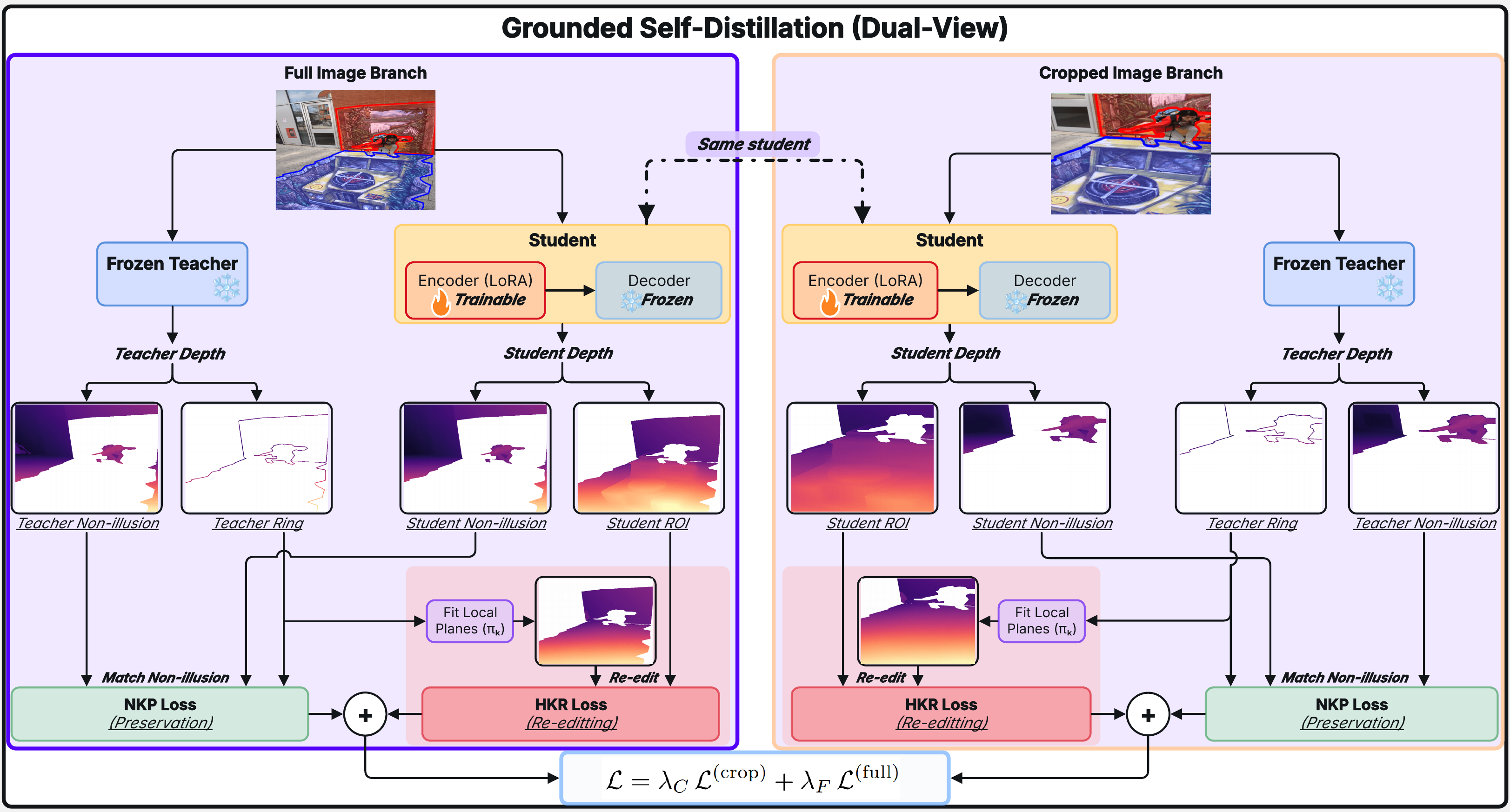}
  \vspace{-6mm}
  \caption{\textbf{Grounded Self-Distillation (Dual-View).}
For each illusion sample, we process both the \textbf{full image} (left) and a \textbf{context-restricted crop} (right). A \textbf{frozen teacher} predicts depth for each view, while a single \textbf{student} (shared weights across both branches) predicts depth using a \textbf{trainable LoRA-augmented encoder} and a \textbf{frozen decoder}. In each branch, we separate the prediction into a \textbf{non-illusion/background region} (and an ROI-adjacent \textbf{ring}) and the \textbf{illusion ROI}. The \textbf{Non-hallucination Knowledge Preservation (NKP)} loss distills the teacher’s stable geometry onto the student on non-illusion/ring regions to prevent catastrophic forgetting. The \textbf{Hallucination Knowledge Re-editing (HKR)} loss suppresses spurious structure inside the ROI by penalizing second-order structure, and re-targeting each ROI instance toward \textbf{local surface hypotheses} fitted from the teacher’s ring neighborhood. The final objective combines both views.}
  \label{fig:pipeline}
  \vspace{-4mm}
\end{figure*}


\section{Methodology: Taming 3D Mirages}
\label{sec:method}

\subsection{Problem Definition}
\label{sec:problem}
Let $f_{\theta}$ be a pre-trained monocular depth estimation foundation model with weights $\theta$.
Given an input image $x \in \mathbb{R}^{H \times W \times 3}$, the model produces a dense depth map $D = f_{\theta}(x)$, where $D \in \mathbb{R}^{H \times W}$.
We define a ``3D Mirage'' as a failure mode characterized by two conditions, identified using a benchmark dataset $\mathcal{D}$.
Each sample in $\mathcal{D}$ consists of a full-context image $x_{\text{full}}$, a context-restricted crop $x_{\text{crop}}$, and a binary mask $m$ defining a Region of Interest (ROI) that serves as a \emph{locally low-curvature carrier} (i.e., physically smooth or piecewise-smooth).

\vspace{1mm}
\noindent\textbf{Failure modes.} We consider the following two failure modes of models:
\begin{enumerate}
    \item \textbf{Geometric Hallucination (Deviation):}
    The model predicts spurious high-curvature 3D structure inside the low-curvature ROI.
    We diagnose this using a fixed second-order operator $\mathcal{L}$ (separable second-difference magnitude), where a geometrically consistent prediction should satisfy
    $\mathcal{L}(D)\odot m \approx 0$.
    A high response indicates a geometric hallucination.

    \item \textbf{Contextual Instability (Confusion):}
    The model's prediction for the \textit{same} physical region changes significantly when surrounding context is altered.
    Let $D_{\text{full}} = f_{\theta}(x_{\text{full}})$ and $D_{\text{crop}} = f_{\theta}(x_{\text{crop}})$.
    From known crop box $b_i$, we resample the full-view depth prediction into the crop coordinate frame and compare it to $D_{\text{crop}}$ on the ROI pixels (mask $M_i$ in the crop frame).
    We call the model contextually unstable when the predicted non-existent 3D structure inside the ROIs differ substantially.
\end{enumerate}

\noindent\textbf{Desired Properties.}
Our goal is to learn a parameter-efficient adaptation $\Delta\theta$ for $f_{\theta}$, producing an adapted model $f_{\theta'}$ with $\theta'=\theta+\Delta\theta$.
Rather than directly optimizing the evaluation metrics, we target the following three \emph{conceptual} properties:
\begin{enumerate}
    \item \textbf{ROI grounding:} suppress spurious curvature inside the ROI, i.e., encourage low $\mathcal{L}(f_{\theta'}(x))$ on pixels in $m$.
    \item \textbf{Context stability:} make the ROI prediction invariant to context removal, so that full-image and aligned crop predictions agree on $m$.
    \item \textbf{Knowledge preservation:} preserve the teacher's behavior outside the ROI to avoid catastrophic forgetting, i.e., keep $f_{\theta'}(x)$ close to $f_{\theta}(x)$ on $(1-m)$.
\end{enumerate}
Our proposed metrics DCS and CCS (Sec.~\ref{sec:metrics}) are reference-free structural \emph{measurement tools} aligned with deviation and confusion. They complement standard metrics and serve as relative probes to use with our benchmark.

\subsection{Grounded Self-Distillation (GSD) Pipeline}

The ROI second-order magnitude-based projection in Sec.~\ref{sec:metrics} isolates two failure modes on illusion ROIs: (i) spurious curvature inside low-curvature carriers (read out radially as DCS) and (ii) context-driven drift between full and crop views (off-diagonal shift as CCS).
We leverage a strong pretrained depth model (Depth-Anything v2) and adapt it so that the network \emph{learns to suppress illusory curvature} and \emph{remains less sensitive to missing context}, while preserving its background/ordinal behavior.
Concretely, the objective mirrors the axes of our evaluation: reduce curvature artifacts inside the ROI and stabilize full/crop predictions without sacrificing non-ROI structure.

We illustrate our whole system pipeline in Fig.~\ref{fig:pipeline}. The 3D Mirage vulnerability largely stems from global context priors in the model's ViT encoder, which we tune directly.
We use LoRA for this adaptation because it surgically modifies encoder behavior in a low-rank subspace, preserving the frozen backbone weights (our ``teacher'') and preventing catastrophic forgetting~\cite{hu2021lora,aghajanyan2021intrinsic}.

Let $T$ be the frozen teacher (\textit{e.g.}, DAv2) and $S$ the student obtained by inserting LoRA adapters into the teacher's \emph{encoder}.
Only LoRA parameters (and a small gating MLP) are trainable; the frozen teacher is used only for training supervision, so inference requires a single student forward pass with the LoRA adapters enabled.
Training is dual-view with shared weights (Fig.~\ref{fig:pipeline}): a crop branch receives $x_{\text{crop}}$ and a full branch receives $x_{\text{full}}$.
Denote student depths by $s$ (crop) and $s^{\!F}$ (full), teacher depths by $t$ and $t^{\!F}$.
Importantly, the re-editing objective does \emph{not} force the illusion to stay on a single flat surface:
we suppress hallucinated second-order structure while allowing piecewise-smooth geometry supported locally via a teacher-guided mixture of simple surface hypotheses and a learned gate (see Fig.~\ref{fig:curvework}).
\vspace{-2mm}
\subsection{Composite Loss Function}
We optimize a weighted sum of terms per branch (crop/full) and then sum branches. 

\noindent\textbf{Notation.}
We use a masked mean operator
$\overline{u}_{\,m} \,=\, \frac{\langle u,m\rangle}{\langle \mathbf{1},m\rangle}$,
with $\langle \cdot,\cdot\rangle$ denoting element-wise inner product over pixels.
We write the magnitude of our fixed separable second-difference operator as $\mathcal{L}(\cdot)$.

\noindent\textbf{Normalization and Masks (Per Branch).}
Each branch is normalized to the teacher's background statistics ($\mu_B,\sigma_B$) computed over a branch-specific background mask $m_{\text{bg}}$ (Eq.~\ref{eq:bgmask}), yielding normalized depths $(z,z_T)$ for the crop view and $(z^{\!F},z_T^{\!F})$ for the full view.

From the binary ROI mask $m$ we form an ROI-adjacent ring $r$ by dilation and subtraction, and a thin guard ring $r_g$ by one additional dilation step.
We then split $r$ into a high-gradient edge subset $r_e$ \emph{(top 10\% by $|\mathcal{L}(z_T)|$ within $r$)} and the complementary low-gradient seam $r_f = r\setminus r_e$.
The background mask excludes the ROI and both rings:
\begin{equation}
m_{\text{bg}}=(1-m)(1-r)(1-r_g),
\label{eq:bgmask}
\end{equation}
(and analogously $m_{\text{bg}}^{F}$ for the full branch).
On the seam $r_f$ we also compute a locally smoothed teacher depth $\tilde z_T$ using ring-restricted local averaging (masked smoothing on $r_f$); we use the analogous $\tilde z_T^{F}$ for the full branch.

Let $\mathcal{R}$ denote the set of ROI instances (polygons) in a given view, with per-instance masks $\{m_j\}_{j\in\mathcal{R}}$ and union mask $m=\bigcup_{j\in\mathcal{R}} m_j$.
Around each ROI instance $j$ we fit up to $K$ simple surface hypotheses $\{\pi_{j,k}\}_{k=1}^{K}$ to the teacher depth on a thin ROI-adjacent band (the ring), and record residual scales $\{\sigma_{j,k}\}$ (lower residual indicates a more plausible local surface explanation).
For the student we define per-instance ROI deviations:
\begin{equation}
\ell_{j,k}=\overline{|\,z-\pi_{j,k}\,|}_{\,m_j}, \qquad k=1,\ldots,K.
\end{equation}
A compact gating network $G$ maps ROI/ring statistics to logits over the $K$ fitted hypotheses.
We set $w_j=\mathrm{softmax}(G(\cdot))$ to get mixture weights $\{w_{j,k}\}_{k=1}^{K}$.
Soft targets $q_j$ are derived from $\{\sigma_{j,k}\}$ (lower residual $\Rightarrow$ higher target weight), and we add a cross-entropy regularizer $\mathrm{CE}(w_j,q_j)$ (with temperature/label-smoothing) and an anchor term $\min_k \ell_{j,k}$.

\noindent\textbf{Hallucination Knowledge Re-editing (HKR) Loss.}
This term directly targets the radial axis (DCS) by collapsing second-order structure inside the ROI toward zero.
We apply a curvature penalty over the union ROI mask $m$, and a teacher-guided mixture loss accumulated over ROI instances $\mathcal{R}$:
\vspace{-4mm}
\begin{equation}
\mathcal{L}_{\text{HKR}} = \alpha_1 \overline{|\mathcal{L}(s)|}_m + \alpha_2 \sum_{j\in\mathcal{R}} \sum_{k=1}^{K} w_{j,k} \ell_{j,k}
\end{equation}
We apply the same HKR form to both crop and full views (with view-specific masks, teacher statistics, and fitted hypotheses), and downweight the full-view contribution in the total objective.
For regularizer images with no illusion ROI annotation, $\mathcal{R}=\varnothing$ and the ROI-local HKR terms are inactive by construction.

\noindent\textbf{Non-hallucination Knowledge Preservation (NKP) Loss.}
This \textbf{\path{self-distillation}} term preserves the teacher's geometry on stable \emph{background} regions using $m_{\text{bg}}$.
To stabilize the transition across the ROI boundary, preserve edge detail and suppress halo artifacts, we use a compact \emph{ring} regularizer that (i) tethers the student's depth to a locally smoothed teacher on the low-gradient seam $r_f$, and (ii) matches second-order structure (via $\mathcal{L}(\cdot)$) on the high-gradient edge subset $r_e$ and the protective guard ring $r_g$.
The resulting loss:
\begin{equation}
\begin{aligned}
\mathcal{L}_{\text{NKP}} ={}& \alpha_3 \overline{|z-z_T|}_{m_{\text{bg}}} + \alpha_4 \overline{|\mathcal{L}(z)-\mathcal{L}(z_T)|}_{m_{\text{bg}}} + \alpha_5 \overline{|z-\tilde z_T|}_{r_f} \\
&+ \alpha_6 \overline{|\mathcal{L}(z)-\mathcal{L}(z_T)|}_{r_e} + \alpha_7 \overline{|\mathcal{L}(z)-\mathcal{L}(z_T)|}_{r_g}
\end{aligned}
\end{equation}
We apply the same NKP structure to the full branch.

\noindent\textbf{Total Objective Loss and Regularization.}
Per branch, the objective is the weighted sum of $\mathcal{L}_{\text{HKR}}+\mathcal{L}_{\text{NKP}}$ plus gating regularizers ($\mathrm{CE}(w,q)$ and the anchor term).
Crop and full branches are combined to bias against context drift while keeping the crop branch dominant:
$\mathcal{L}=\lambda_C\,\mathcal{L}_{\text{crop}}+\lambda_F\,\mathcal{L}_{\text{full}}$.
The full branch uses the same HKR/NKP structure.

To avoid degenerate over-smoothing, illusion batches are interleaved with non-illusion data during the optimization steps. We incorporate two real-image datasets during training:
The Penn--Fudan dataset provides 170 urban street images with 345 upright pedestrians, offering diverse occlusions and pedestrian scales~\cite{pennfudanped}.
The CamVid collection contributes 701 raw still frames of urban driving scenes widely used in autonomous-driving research~\cite{brostow2009camvid}. For these regularizer images, training reduces to the preservation objective for the scenes. 
This regularization helps suppress over-flattening and edge drift without weakening training supervision, and targets safety-critical deployments of depth models.


\vspace{-2mm}

\section{Experiments}
\label{sec:experiments}
\vspace{-2mm}
To validate our framework, we first establish the vulnerability of SOTA models on our \textbf{3D-Mirage} benchmark. We then demonstrate the effectiveness of our \textbf{Grounded Self-Distillation} method in taming these hallucinations.
\vspace{-2mm}
\subsection{Experimental Setup}
\vspace{-2mm}

\subsubsection{Baselines}
We compare against a comprehensive suite of SOTA monocular depth foundation models using their official weights. This includes the \textbf{Depth Anything families} (DA-\{S,B,L\} \cite{dav1} and DAv2-\{S,B,L\} \cite{dav2}, including the indoor (DAv2-I) and outdoor (DAv2-O) specialized variants) and \textbf{other foundation models} (DepthPro \cite{Bochkovskii_2025_ICLR}, Marigold \cite{Ke_2024_CVPR}, DepthFM \cite{Gui_2025_AAAI}, ZoeDepth \cite{zoe}, and MiDaS \cite{ranftl2022tpami}).

Our primary baseline for adaptation is \textbf{\path{Depth-Anything-V2-Large-hf} (DAv2-L)} \cite{dav2}, which serves as the frozen \textbf{teacher model ($T$)} and the initial backbone for our \textbf{student model ($S$)}.

\vspace{-2mm}

\begin{wraptable}{r}{0.65\textwidth}
\vspace{-10mm}
\centering
\caption{\textbf{Quantitative comparison on the 3D-Mirage benchmark.} We evaluate SOTA foundation models and our method (Grounded Self-Distillation) using our proposed metrics. \textbf{Lower is better.} Our method (Ours) drastically reduces both geometric deviation (DCS) and contextual instability (CCS) compared to all baselines, including its own teacher model (DAv2-L). $\Delta$ denotes the relative improvement of our model over the DAv2-L baseline.}
\label{tab:illusion}
\resizebox{\linewidth}{!}{%
\begin{tabular}{l | c c c | c c c}
\toprule
\textbf{Model} &
$d_{\textbf{cluster}}$$\downarrow$ & $d_{\textbf{avg}}$$\downarrow$ & \textbf{DCS}$\downarrow$ &
$D_{\textbf{cluster}}$$\downarrow$ & $D_{\textbf{avg}}$$\downarrow$ & \textbf{CCS}$\downarrow$ \\
\midrule
DepthPro~\cite{Bochkovskii_2025_ICLR} &  {317.8} &  {331.4} &  {649.1} &  {6.680e-4} &  {9.290e-4} &  {1.597e-3} \\
Marigold~\cite{Ke_2024_CVPR} &  {701.1} &  {726.2} &  {1.427e3} &  {2.294e-3} &  {2.402e-3} &  {4.696e-3} \\
DepthFM~\cite{Gui_2025_AAAI} &  {1.020e3} &  {1.063e3} &  {2.083e3} &  {4.914e-3} &  {5.215e-3} &  {1.013e-2} \\
ZoeDepth~\cite{zoe} &  {291.5} &  {297.8} &  {589.3} &  {7.486e-4} &  {7.560e-4} &  {1.505e-3} \\
MiDaS~\cite{ranftl2022tpami} &  {330.2} &  {340.0} &  {670.2} &  {4.120e-4} &  {5.090e-4} &  {9.220e-4} \\
\midrule
DA-S~\cite{dav1}  &  {225.9} &  {233.9} &  {459.8} &  {3.190e-4} &  {3.570e-4} &  {6.760e-4} \\
DA-B~\cite{dav1}  &  {236.0} &  {246.7} &  {482.7} &  {2.710e-4} &  {3.520e-4} &  {6.230e-4} \\
DA-L~\cite{dav1} &  {243.3} &  {251.7} &  {495.0} &  {2.730e-4} &  {3.290e-4} &  {6.030e-4} \\ \midrule
DAv2-IS~\cite{dav2} &  {415.6} &  {424.5} &  {840.1} &  {1.452e-3} &  {1.473e-3} &  {2.924e-3} \\
DAv2-IB~\cite{dav2} &  {347.5} &  {359.5} &  {706.9} &  {1.133e-3} &  {1.183e-3} &  {2.315e-3} \\
DAv2-IL~\cite{dav2} &  {406.1} &  {418.4} &  {824.5} &  {1.161e-3} &  {1.187e-3} &  {2.348e-3} \\
DAv2-OS~\cite{dav2} &  {685.4} &  {698.4} &  {1.384e3} &  {3.101e-3} &  {3.102e-3} &  {6.203e-3} \\
DAv2-OB~\cite{dav2} &  {713.4} &  {726.1} &  {1.439e3} &  {2.901e-3} &  {2.902e-3} &  {5.804e-3} \\
DAv2-OL~\cite{dav2} &  {537.0} &  {547.5} &  {1.085e3} &  {1.959e-3} &  {1.961e-3} &  {3.920e-3} \\
DAv2-S~\cite{dav2} &  {495.9} &  {511.2} &  {1.007e3} &  {7.210e-4} &  {7.900e-4} &  {1.512e-3} \\
DAv2-B~\cite{dav2} &  {431.7} &  {449.3} &  {881.0} &  {6.320e-4} &  {7.270e-4} &  {1.359e-3} \\
DAv2-L~\cite{dav2} (\textbf{Baseline}) &  {488.8} &  {505.8} &  {994.6} &  {6.840e-4} &  {7.820e-4} &  {1.466e-3} \\
\midrule
\textbf{Ours} &
\textbf{ {28.55}} & \textbf{ {30.09}} & \textbf{ {58.64}} &
\textbf{ {9.174e-5}} & \textbf{ {9.894e-5}} & \textbf{ {1.907e-4}} \\
\midrule
\textbf{$\Delta$ (\%)} &
(\textbf{-94.16}\%) & (\textbf{-94.05}\%) & (\textbf{-94.10}\%) &
(\textbf{-86.59}\%) & (\textbf{-87.35}\%) & (\textbf{-86.99}\%) \\
\bottomrule
\end{tabular}
}
\vspace{-6mm}
\end{wraptable}
\subsubsection{Implementation Details}
We implement our method in PyTorch, using the PEFT library \cite{houlsby2019parameter} for LoRA adaptation.

\noindent\textbf{1) Data.} We use a custom sampler with a 4:1 ratio of 3D-Mirage (positive) samples to regularizer (negative) samples. Negative samples are drawn from Penn-Fudan \cite{pennfudanped} and CamVid \cite{brostow2009camvid} to prevent catastrophic forgetting/flattening on standard street scenes. We apply 50\% horizontal flip and 5\% photometric jitter augmentations. We split the benchmark at the \emph{source-image} level to avoid leakage between a full image and its context-restricted crops. All crops derived from the same original image are assigned to the same split. We use a 90/10 train/test split, yielding 421/47 scenes and 1684/188 paired full-crop instances.

\noindent\textbf{2) Model.} We inject LoRA adapters (rank $r=16$, $\alpha=32$, dropout $0.05$, `bias=none`) into the DINOv2 encoder's patch embedding layer and all MLP linear layers (`fc1`, `fc2`) within the 24 transformer blocks. This results in only \textbf{4M trainable parameters} ($\approx$1.2\% of the DAv2-L backbone).

\noindent\textbf{3) Training.} We use the AdamW optimizer with a learning rate of $1 \times 10^{-4}$, weight decay of 0.01, and global gradient clipping of 1.0. For training stability, all student and teacher depth outputs are z-normalized over background pixels before loss computation. The model is trained for only \textbf{1 epoch} with a batch size of 8 on an NVIDIA A100 GPU. Extending training to 4 epochs yields only marginal gains on DCS/CCS, while the other evaluation metrics show mixed behavior. Detailed results are provided in the supplementary.

For the losses, we fix the loss weights to $\alpha_1{=}1.0$, $\alpha_2{=}0.4$, $\alpha_3{=}1.0$, $\alpha_4{=}0.5$, $\alpha_5{=}0.3$, $\alpha_6{=}0.8$, and $\alpha_7{=}0.3$ to keep the different losses numerically comparable.

\vspace{-2mm}
\subsubsection{Evaluation}
We evaluate models on two fronts. First, we test for \textbf{hallucination robustness} using our \textbf{3D-Mirage} benchmark with the proposed \textbf{DCS} (hallucination intensity) and \textbf{CCS} (contextual instability) metrics, where lower is better. Second, we test for \textbf{knowledge preservation} using an ordinal pairwise accuracy protocol on \textbf{NYU-v2} \cite{nyu} (as detailed in Sec. \ref{sec:ablation_setup}) to ensure our method does not catastrophically forget general depth estimation.

\subsection{Main Results: Taming 3D Mirages}

\begin{wrapfigure}{r}{0.5\textwidth}
\vspace{-6mm}
    \centering
    \includegraphics[width=0.95\linewidth]{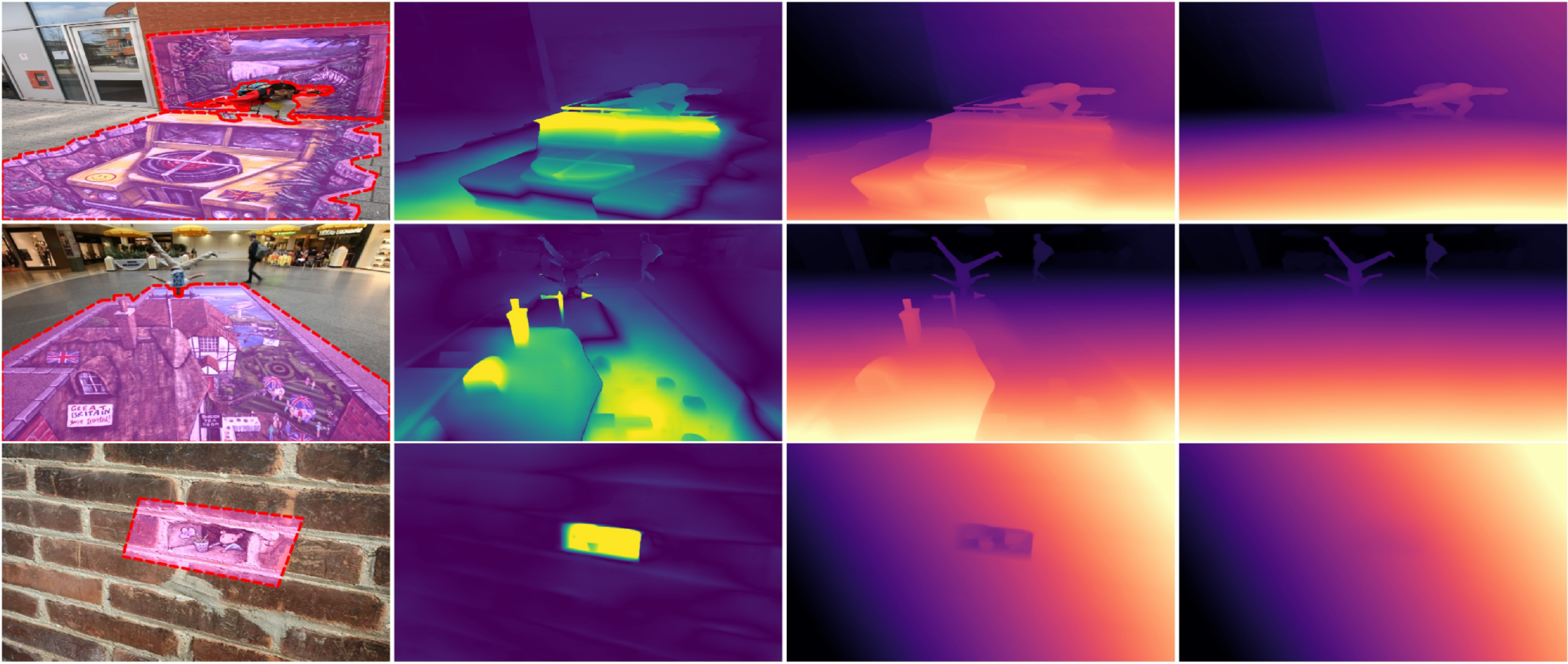}\vspace{-2mm}
    \caption{\textbf{Qualitative results of our Grounded Self-Distillation.} Each row compares our model to the baseline on a 3D-Mirage sample. (1) Input RGB. (2) Error heatmap (Ours vs. Baseline), showing changes are confined to the ROI. (3) Baseline (DAv2-L) depth, which hallucinates non-existent 3D structures. (4) Our model's depth, which correctly perceives the planar surface. Our method tames the 3D mirage without distorting the background, including illusions existing on separate planes (first row).}
    \label{fig:sample}
\vspace{-8mm}
\end{wrapfigure}
Table \ref{tab:illusion} presents the quantitative results on our 3D-Mirage benchmark. The results are decisive: \textbf{Across all evaluated SOTA foundation models, we observe consistently elevated DCS/CCS on 3D-Mirage, suggesting these models are highly vulnerable to 3D mirages.} The failure is systemic, afflicting all tested architectures (transformer, diffusion-based, etc.). This suggests that their training condition has inadvertently created powerful, dataset-level priors (e.g., complex 2D patterns often imply high second-order 3D structure) that override local geometric evidence when faced with ambiguous, out-of-distribution perceptual traps. Our baseline, DAv2-L, scores a high 994.6 on DCS, confirming it perceives significant, spurious 3D geometry.

In contrast, our GSD method achieves a DCS of only \textbf{58.64} and a CCS of \textbf{1.907e-4}, the best scores by a large margin. This represents a massive \textbf{94\% reduction in geometric deviation (DCS)} and an \textbf{87\% reduction in contextual instability (CCS)} compared to the DAv2-L teacher. This demonstrates not only that the hallucination is removed, but that the model is much less confused by the removal of context. It has learned to ground its prediction in local geometric evidence (low-curvature ROI) rather than being swayed by fragile semantic priors.

\begin{wrapfigure}{r}{0.5\textwidth}
\vspace{-6mm}
    \centering
    \includegraphics[width=0.95\linewidth]{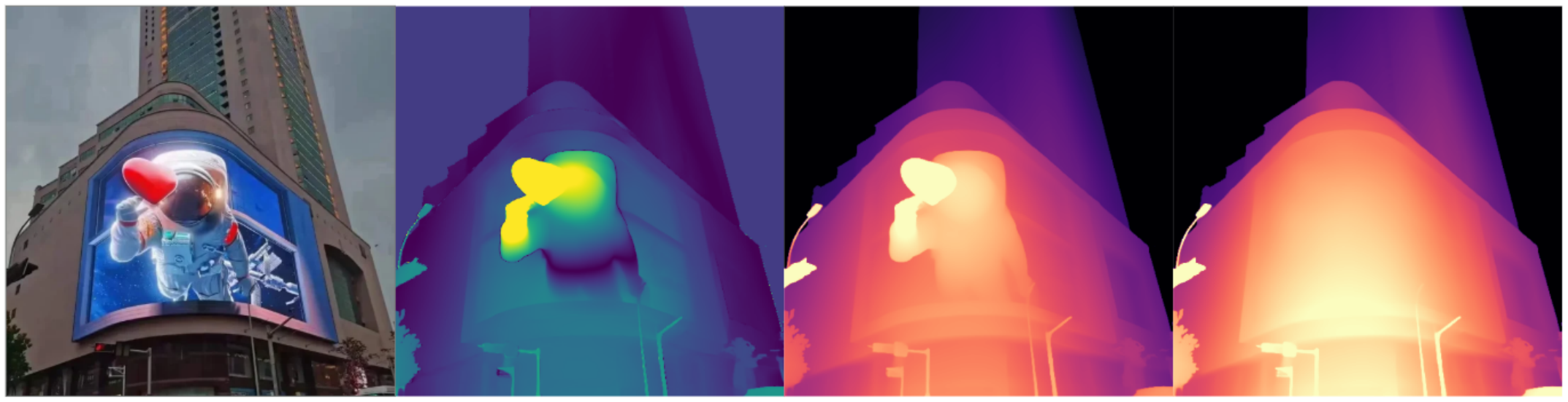}\vspace{-2mm}
    \caption{\textbf{Impact of plane-mixture and gating}. Our HKR loss does not force a single fronto-parallel plane. Our model learns to approximate curvature and can partially recover curved surface from hallucination.}
    \label{fig:curvework}
    \vspace{-8mm}
\end{wrapfigure}
This illusion taming is visualized in Fig.~\ref{fig:sample}, ~\ref{fig:curvework}. Our model (column 4) successfully identifies and resolve hallucination, including cases on multiple surfaces, and, in some cases, curved surfaces. The baseline (column 3) dangerously hallucinates large obstacles, caverns, and non-existent 3D structure. Crucially, the difference heatmap (column 2) confirms that our model's corrections are surgically confined to the illusion ROI. This provides strong evidence that our $\mathcal{L}_{\text{NKP}}$ (knowledge preservation) loss is working as intended, preventing the flattening objective from leaking and destroying valid geometry in the background.

\vspace{-2mm}
\paragraph{Generalization to 3D visual illusions (3DVI).}
\begin{wraptable}{r}{0.65\textwidth}
\vspace{-10mm}
\centering
\caption{\textbf{Results on illusion regions of the 3D-Visual-Illusion (3DVI) test set.} ``align'' denotes globally shared affine (scale+shift) alignment computed from ground truth, following Yao et al.\ (NeurIPS'25).}
\scriptsize
\setlength{\tabcolsep}{3pt}
\renewcommand{\arraystretch}{1.05}
\resizebox{\linewidth}{!}{%
\begin{tabular}{l c r r r r r r r}
\toprule
\multirow{2}{*}{Method} & \multirow{2}{*}{FT} &
\multicolumn{4}{c}{Disparity Space} & \multicolumn{3}{c}{Depth Space} \\
\cmidrule(lr){3-6}\cmidrule(lr){7-9}
& & EPE$\downarrow$ & bad2$\downarrow$ & bad3$\downarrow$ & bad5$\downarrow$ &
AbsRel$\downarrow$ & RMSE$\downarrow$ & $\delta_1\uparrow$ \\
\midrule
\multicolumn{9}{l}{\textit{Stereo or multi-view input models (not our focus)}} \\
Dust3R~\cite{dust3r_cvpr2024}                & $\times$      & 6.74 & 52.89 & 45.31 & 36.61 & 0.25 & 0.22 & 87.09 \\
VGGT~\cite{vggt_cvpr2025}                 & $\times$      & 6.16 & 53.32 & 44.89 & 37.20 & 0.13 & 0.12 & 78.46 \\
RAFT-Stereo~\cite{raft_stereo_3dv2021}           & $\times$      & 1.62 & 24.32 & 13.20 & 2.97  & 0.04 & 0.06 & 99.18 \\
Selective-RAFT~\cite{selective_stereo_cvpr2024}         & $\times$      & 1.58 & 23.46 & 12.65 & 2.57  & 0.03 & 0.07 & 99.60 \\
Selective-IGEV~\cite{igev_stereo_cvpr2023}        & $\times$      & 1.67 & 24.06 & 13.11 & 2.99  & 0.04 & 0.10 & 99.26 \\
MochaStereo~\cite{mocha_stereo_cvpr2024}           & $\times$      & 1.75 & 25.49 & 14.11 & 3.54  & 0.04 & 0.11 & 98.76 \\
StereoAnything~\cite{stereoanything}        & $\times$      & 2.41 & 29.00 & 16.15 & 6.54  & 0.11 & 0.32 & 96.23 \\
\textbf{3DVI}~\cite{3dvi}     & $\checkmark$  & \textbf{1.77} & \textbf{26.72} & \textbf{15.73} & \textbf{3.60}  & \textbf{0.03} & \textbf{0.08} & \textbf{99.60} \\
\midrule
\multicolumn{9}{l}{\textit{Monocular or single-view input models (our focus)}} \\
DAv2~\cite{dav2}                  & $\times$ & 5.81 & 61.45 & 43.18 & 30.57 & 0.14 & 0.15 & 92.86 \\
Metric3D~\cite{metric3d_iccv2023}              & $\times$ & 12.46 & 94.11 & 91.14 & 82.05 & 0.34 & 0.29 & 48.97 \\
DAv2 metric~\cite{dav2}           & $\times$ & 16.24 & 92.53 & 87.43 & 75.25 & 0.52 & 0.39 & 48.75 \\
DepthPro~\cite{depth_pro_2024}              & $\times$ & 12.26 & 87.08 & 80.60 & 62.43 & 0.28 & 0.25 & 65.92 \\
Marigold~\cite{Ke_2024_CVPR}              & $\times$ & 21.16 & 65.67 & 59.67 & 53.19 & 0.45 & 0.37 & 63.65 \\
DAv2 metric(align)   & $\times$ & 5.23 & 56.82 & 45.50 & 28.89 & 0.17 & 0.15 & 93.70 \\
Metric3D(align)      & $\times$ & 5.70 & 66.26 & 50.92 & 40.43 & 0.17 & 0.17 & 94.80 \\
DepthPro(align)      & $\times$ & 4.36 & 44.98 & 34.98 & 24.70 & 0.09 & 0.10 & 93.83 \\
\textbf{Ours}         & $\times$ & \textbf{1.75} & \textbf{26.67} & \textbf{15.52} & \textbf{6.59} & \textbf{0.03} & \textbf{0.06} & \textbf{99.50} \\
\bottomrule
\end{tabular}%
}
\label{tab:3dvi_r1}
\vspace{-4mm}
\end{wraptable}
We additionally evaluate our model zero-shot on the real-world test split of the 3D-Visual-Illusion~\cite{3dvi,3dvi_code}  (3DVI) dataset from Yao et al.\ (NeurIPS'25), following their evaluation protocol (Table~\ref{tab:3dvi_r1}), which reports disparity-space metrics (EPE, bad-$t$) and depth-space metrics (AbsRel, RMSE, $\delta_1$) over the provided illusion masks under a single global scale-and-shift alignment.

\noindent\textbf{Cross-architecture generalization.} The GSD objective is not tied to the DAv2 encoder. Applying the same recipe to a second transformer backbone (ZoeDepth) and to a diffusion backbone (Marigold) yields consistent reductions on both 3D-Mirage and 3DVI: on ZoeDepth (Table~\ref{tab:zoedepth_transfer}), DCS drops $589\rightarrow335$; on Marigold (Fig.~\ref{fig:marigold_transfer}), a LoRA ($r{=}16$) adaptation cuts DCS and CCS by \textbf{44\%} and \textbf{40\%} while preserving DA-2K/DIW depth, with full results in the supplement.

\begin{center}
{\scriptsize
\setlength{\tabcolsep}{4pt}
\renewcommand{\arraystretch}{1.05}
\vspace{-2mm}
\captionof{table}{\textbf{Transfer to ZoeDepth} (a second transformer backbone): 3D-Mirage (DCS/CCS) and 3DVI metrics. Lower is better except $\delta_1$.}
\resizebox{0.92\linewidth}{!}{%
\begin{tabular}{l cc ccccc}
\toprule
\textbf{Model} & \textbf{DCS}$\downarrow$ & \textbf{CCS}$\downarrow$ & \textbf{EPE}$\downarrow$ & \textbf{bad2}$\downarrow$ & \textbf{AbsRel}$\downarrow$ & \textbf{RMSE}$\downarrow$ & $\boldsymbol{\delta_1}\uparrow$ \\
\midrule
ZoeDepth & 589.27 & 1.505e-3 & 9.555 & 67.66 & 0.1899 & 0.2255 & 76.16 \\
\textbf{Our ZD} & \textbf{335.12} & \textbf{1.136e-3} & \textbf{7.286} & \textbf{61.70} & \textbf{0.1273} & \textbf{0.1400} & \textbf{84.76} \\
\bottomrule
\end{tabular}}}
\vspace{3mm}

\label{tab:zoedepth_transfer}
    \includegraphics[width=0.62\linewidth]{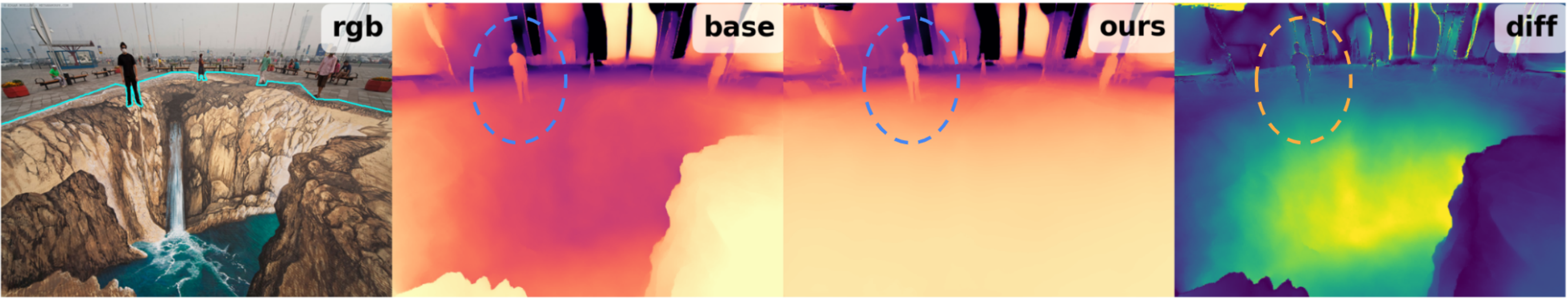}
    \captionof{figure}{\textbf{Qualitative depth on the Marigold diffusion backbone} (from left: RGB, baseline, ours, difference). GSD removes the hallucinated 3D structure on the illusion surface (dashed) while preserving fine detail such as the standing person.}
    \label{fig:marigold_transfer}
\end{center}

\subsection{Ablation Study}
\label{sec:ablation_setup}
\begin{figure*}[tb]
\centering
\includegraphics[width=0.47\linewidth]{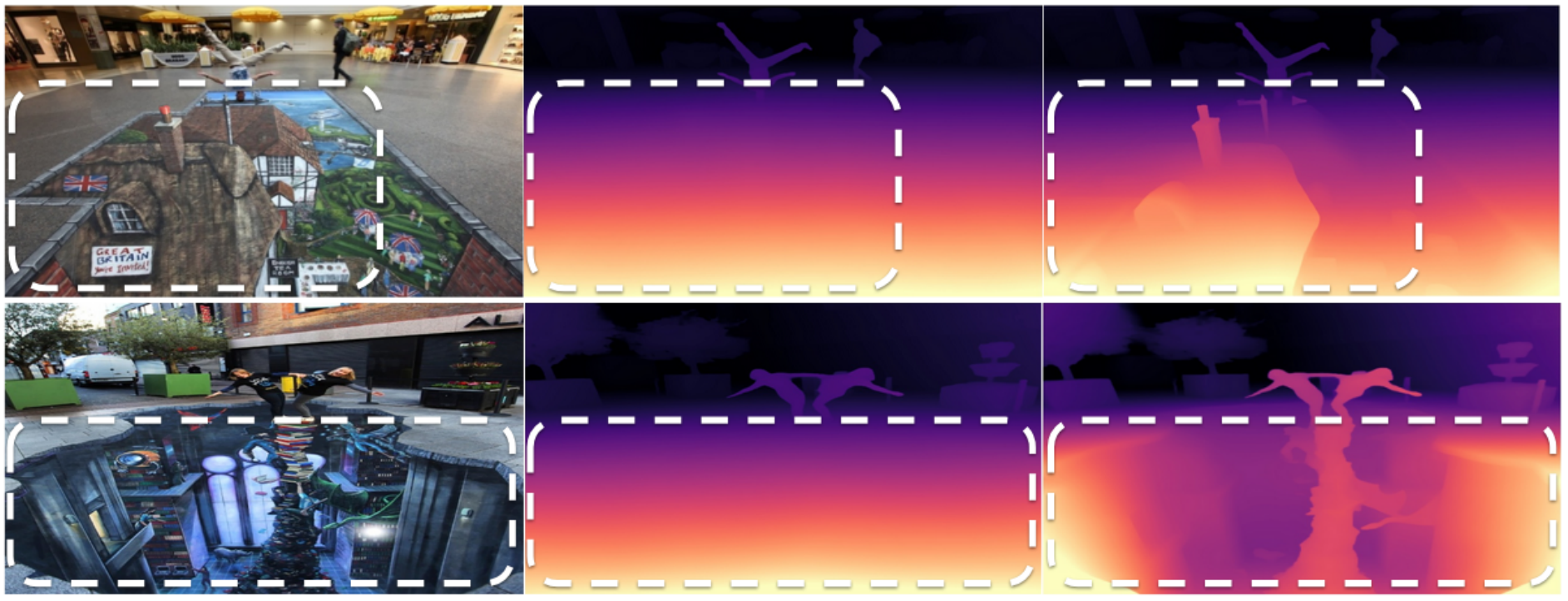}\hfill
\includegraphics[width=0.47\linewidth]{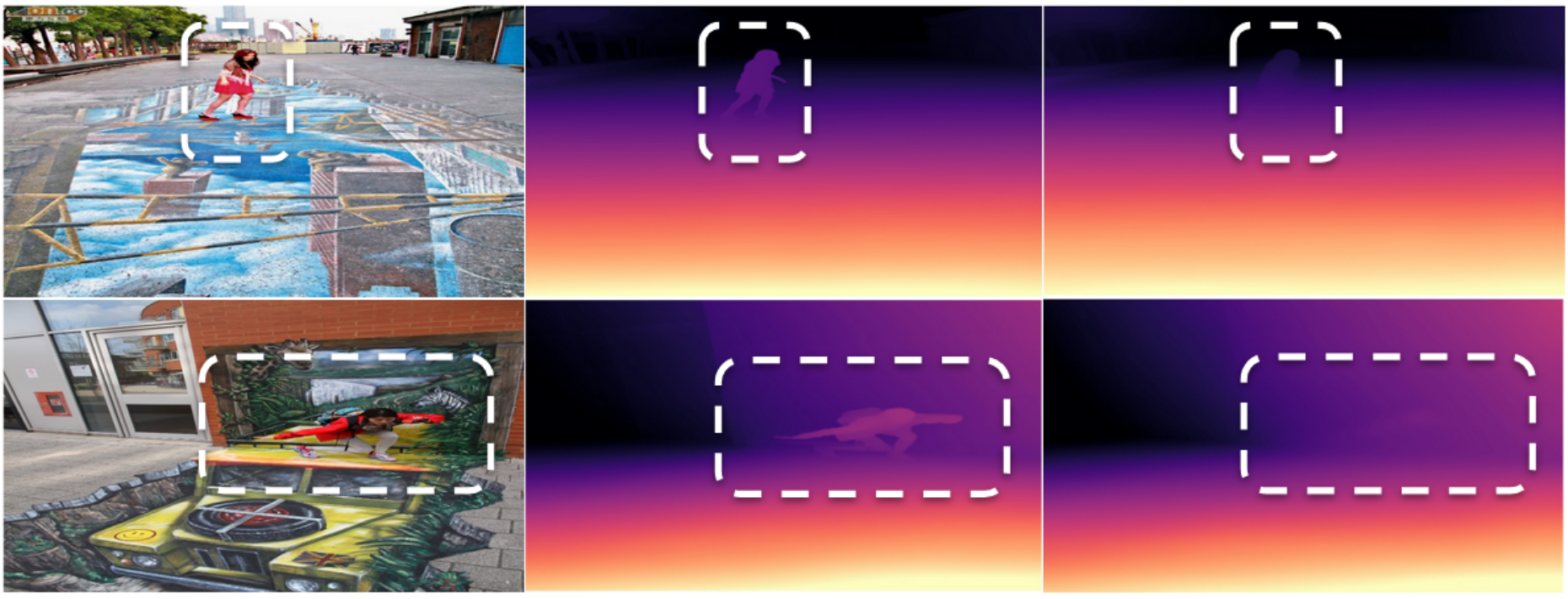}
\vspace{-2mm}
\caption{\textbf{Qualitative ablation of the two loss terms} (each panel: input RGB, our full model, ablated variant). \emph{Left} (w/o $\mathcal{L}_{\text{HKR}}$): the background is preserved but the 3D mirage on the road is left intact. \emph{Right} (w/o $\mathcal{L}_{\text{NKP}}$): the mirage is removed but the flattening leaks into the background and distorts real objects. This mirrors the quantitative trade-off in Table~\ref{tab:reldepth}.}
\label{fig:abl_qual}
\vspace{-2mm}
\end{figure*}

\begin{wraptable}{r}{0.5\textwidth}
\vspace{-8mm}
\centering
\caption{\textbf{Component ablation.}}
\resizebox{\linewidth}{!}{%
\begin{tabular}{lccc}
\toprule
\textbf{Benchmark: Metric} &
\textbf{\begin{tabular}[c]{@{}c@{}}No Halluc.\\ Re-editing\end{tabular}} &
\textbf{\begin{tabular}[c]{@{}c@{}}No Knowl.\\ Preserv.\end{tabular}} &
\textbf{Ours} \\
\midrule
DCS$\downarrow$ & 988.60 & 42.83 & 58.64 \\
CCS$\downarrow$ ($\times 10^{-3}$) & 1.470 & 0.141 & 0.191 \\
\midrule
NYUv2 AbsRel$\downarrow$ & 0.1623 & 0.1533 & 0.1597 \\
NYUv2 $\delta_1$ (\%)$\uparrow$ & 79.15 & 80.25 & 79.58 \\
\midrule
KITTI15~\cite{menze2015object} AbsRel$\downarrow$ & 0.3409 & 0.3510 & 0.3424 \\
KITTI15~\cite{menze2015object} $\delta_1$ (\%)$\uparrow$ & 39.33 & 38.27 & 39.21 \\
\midrule
DA-2K: Rel. Acc. (\%)$\uparrow$ & 97.05 & 94.00 & 96.08 \\
\bottomrule
\end{tabular}}
\label{tab:reldepth}
\vspace{-6mm}
\end{wraptable}
We ablate both terms of our objective: \textit{Hallucination Knowledge Re-editing} ($\mathcal{L}_{\text{HKR}}$) suppresses the mirage, and \textit{Non-hallucination Knowledge Preservation} ($\mathcal{L}_{\text{NKP}}$) prevents over-flattening and preserves general depth. We compare the full model against two loss-removal variants on 3D-Mirage (DCS/CCS), NYUv2, KITTI~15, and DA-2K~\cite{dav2} (Table~\ref{tab:reldepth}). Removing $\mathcal{L}_{\text{HKR}}$ preserves standard depth but leaves the 3D mirage largely intact; removing $\mathcal{L}_{\text{NKP}}$ reaches the lowest DCS/CCS but degrades the standard metrics, indicating over re-editing. Our full model gives the best overall trade-off; the qualitative effect of each term is shown in Fig.~\ref{fig:abl_qual}.

We further compare GSD against simpler adaptation strategies (Table~\ref{tab:ablation_simp}): a direct ROI \textit{Simple L1}, naive encoder finetuning (\textit{Ftune Enc.}), and decoder finetuning (\textit{Ftune Dec.}). The Simple L1 variant trains the LoRA student with only an ROI L1 objective pulling the prediction toward a plane fitted from the frozen teacher's ROI-adjacent ring. Our full pipeline gives the best balance of illusion robustness and general-depth performance, while encoder finetuning degrades generalization most and Simple L1 yields smaller gains.

\begin{table*}[t]
\vspace{-2mm}
\centering
\caption{\textbf{Ablation: our full GSD pipeline vs simpler adaptation strategies.} Lower is better except DA-2K, $\delta_1$, and BG~$R^2$.}
\vspace{-2mm}
\scriptsize
\setlength{\tabcolsep}{2pt}
\renewcommand{\arraystretch}{1.05}
\resizebox{\linewidth}{!}{%
\begin{tabular}{l cc|c|c|cccc|ccc}
\hline
\textbf{Model} &
\textbf{DIW (\%)} &
\textbf{DA-2K (\%)} &
\textbf{NYUv2} &
\textbf{KITTI15} &
\multicolumn{4}{c|}{\textbf{3DVI}} &
\multicolumn{3}{c}{\textbf{3D-Mirage}} \\
\cline{4-4} \cline{5-5} \cline{6-9} \cline{10-12}
&
WHDR$\downarrow$ &
pairwise$\uparrow$ &
\textbf{RMSE} $\downarrow$ &
\textbf{RMSE} $\downarrow$ &
\textbf{EPE} $\downarrow$ &
\textbf{bad2} $\downarrow$ &
\textbf{RMSE} $\downarrow$ &
$\boldsymbol{\delta_1}$ \textbf{(\%)} $\uparrow$ &
\textbf{BG $R^2$} $\uparrow$ &
\textbf{DCS} $\downarrow$ &
\textbf{CCS} $\downarrow$ \\
\hline
Ours &
11.48 & 96.08 &
0.5239 &
6.85 &
1.750 & 26.67 & 0.06383 & 99.50 &
0.9395 & 58.64 & 1.907e-4 \\
Simple L1 &
12.03 & 93.04 &
0.5299 &
7.10 &
2.055 & 32.80 & 0.06970 & 98.77 &
0.8965 & 59.68 & 2.118e-4 \\
Ftune Enc. &
32.15 & 58.51 &
0.9014 &
8.54 &
7.629 & 69.93 & 0.1787 & 81.64 &
0.6296 & 27.85 & 9.389e-5 \\
Ftune Dec. &
14.46 & 87.48 &
0.6487 &
6.93 &
4.430 & 50.93 & 0.1137 & 93.15 &
0.8992 & 70.26 & 1.986e-4 \\
\hline
\end{tabular}}
\label{tab:ablation_simp}
\vspace{-4mm}
\end{table*}

\section{Conclusion}
\label{sec:conclusion}
\vspace{-2mm}
We identified, diagnosed, and mitigated a critical vulnerability in SOTA monocular depth models: the \textbf{3D Mirage}, a systemic failure where models hallucinate spurious 3D structure from ambiguous texture, posing a significant risk to safety-critical applications. To our knowledge, this is the first end-to-end framework to \textbf{probe} the failure with our GT-free \textbf{3D-Mirage} benchmark, \textbf{score} it with novel second-order magnitude-based metrics, \textbf{DCS} (structural deviation) and \textbf{CCS} (contextual stability), and \textbf{tame} it with a parameter-efficient \textbf{Grounded Self-Distillation} strategy. Guided by a composite $\mathcal{L}_{\text{HKR}}$ (re-editing) and $\mathcal{L}_{\text{NKP}}$ (preservation) loss, our method reduces hallucinations by \textbf{over 94\%} and instability by \textbf{87\%}, and our ablations confirm the adaptation is surgically precise, avoiding the catastrophic forgetting of naive finetuning. This work provides the essential tools to advance MDE from simple pixel accuracy toward the structural and contextual robustness required for real-world deployment.

\vspace{1mm}
\noindent\textbf{Limitations and Future Work.}
\textcolor{black}{ Our \textbf{3D-Mirage} benchmark is primarily focused on low-curvature surfaces perceived as 3D structure, which does not encompass the full spectrum of perceptual ambiguity, such as texture-less surfaces, reflections, glass, shadows, or adverse weather. Our LoRA-based mitigation was demonstrated on a transformer-based MDE architecture; we further validate its transfer to a second transformer backbone and a diffusion model (Marigold) in the supplement. We do not yet explicitly evaluate on real road hazards such as potholes, curbs, or speed bumps, which would be a valuable safety-oriented complement to our current preservation benchmarks.}

\enlargethispage{5\baselineskip}
\vspace{1mm}
\noindent\textbf{Broader Impact.} Monocular depth models are increasingly deployed in safety-critical perception, where a hallucinated bump on a flat road or wall can trigger unnecessary braking or unsafe maneuvers. Our contributions target this risk end-to-end: the \textbf{3D-Mirage} benchmark exposes the failure, the reference-free \textbf{DCS}/\textbf{CCS} metrics let practitioners audit for spurious curvature without ground-truth depth (rarely available at deployment scale), and \textbf{Grounded Self-Distillation} mitigates it with an adapter that folds into the backbone at inference with no added latency or memory. Removing hallucinated structure must not be mistaken for a guarantee of metric depth accuracy: DCS and CCS certify structural and contextual stability on the low-curvature carrier, not absolute scale, and should complement rather than replace task-level validation before deployment.

\clearpage

\setcounter{page}{1}
\appendix

\vspace{0.5em}

\renewcommand{\thesection}{\Alph{section}}
\renewcommand{\thefigure}{\Alph{figure}}
\renewcommand{\thetable}{\Alph{table}}
\setcounter{figure}{0}
\setcounter{table}{0}
\setcounter{section}{0}

\section{Visualization of Ring, Edge, Seam, and Guard}
\label{subsec:vis-rgs}
\begin{wrapfigure}{r}{0.5\textwidth}
\vspace{-8mm}
    \centering
    \includegraphics[width=0.98\linewidth]{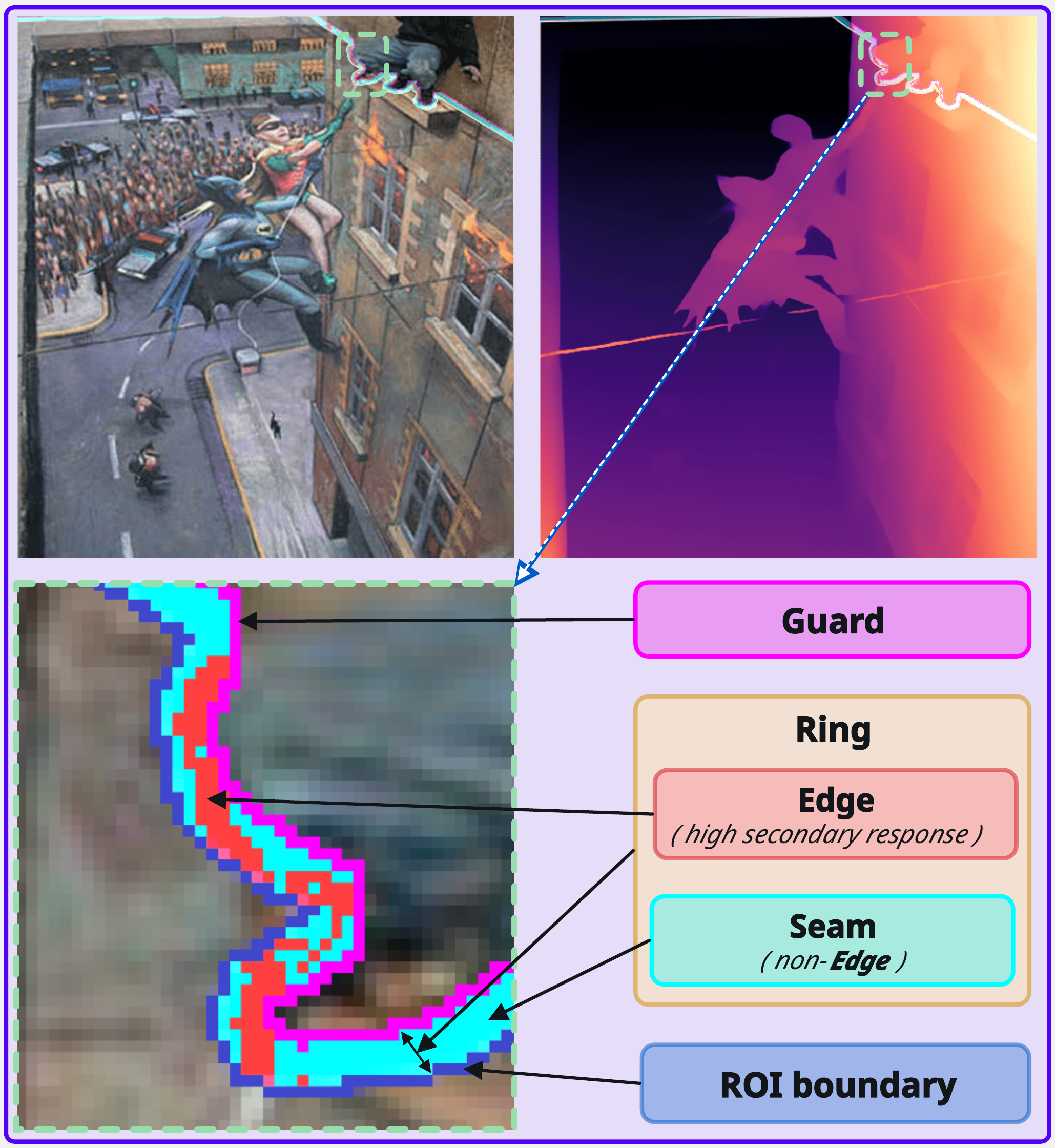}\vspace{-2mm}
\caption{\textbf{Ring, Edge, Seam, and Guard}. The \textbf{ring} is the first dilated band from ROI boundary, and \textbf{guard}(\textbf{magenta}) is the outermost. \textbf{Ring} is split into \textbf{edge}(\textit{red}) - high second-order teacher response - and \textbf{seam}(\textit{cyan}) - smoother remainder. In this example, the man's knee lies inside the ring, its edge is correctly marked as red. Local ring neighborhood is also used for plane-mixture fit.}
    \label{fig:resg}
    \vspace{-6mm}
\end{wrapfigure}
\paragraph{Here we further discuss implementation details of the boundary rings to manage geometric transitions.}
To ensure a stable geometric transition between the re-edited illusion ROI and the reliable surrounding scene, we introduce a structured boundary immediately outside the Region of Interest (ROI). In both the crop and full branches, this boundary stabilizes the transition around the ROI, while the ROI itself is re-edited and the surrounding scene is regularized toward the frozen teacher. This \textbf{\textit{boundary}} region consists of four specific components, as visualized in Fig.~\ref{fig:resg}. First, the primary ROI-adjacent \textbf{ring} $r$ is constructed by applying a 3-pixel dilation to the illusion mask $m$ and subtracting the original ROI, forming the first narrow band. Next, an outer \textbf{guard} ring $r_g$ is formed by one additional 1-pixel dilation step beyond $r$, providing a thin protective band just outside the primary ring.

\paragraph{Preserving stable scene geometry outside the illusion.}
Background preservation is enforced through self-distillation losses on areas entirely outside the illusion and its immediate boundaries. Specifically, the background mask is defined as:
\[
m_{\mathrm{bg}} = (1-m)(1-r)(1-r_g)
\]
While this background is regularized globally toward the teacher, the ring and guard regions are handled separately to stabilize the transition near the ROI boundary.

\paragraph{Preserving real edges while smoothing the illusion transition.}
To balance sharpness and smoothness, the primary ring $r$ is further subdivided into an \textbf{edge} subset $r_e$ and a complementary \textbf{seam} subset $r_f$ using the frozen teacher's branch-normalized depth $z_T$. To achieve this, the code computes the second-order response magnitude $|\mathcal{L}(z_T)|$ and calculates the $90$th percentile across all ring pixels in the branch. The edge subset $r_e$ is defined as the ring pixels exceeding this threshold, representing the sharper, high-response features. The seam subset $r_f = r \setminus r_e$ contains the smoother remainder of the ring.

\paragraph{Enforcing structural agreement and suppressing artifacts.}
Each of these subdivided boundary components serves a distinct functional role. The seam is used to build a masked, locally smoothed teacher target $\tilde{z}_T$ via normalized local averaging, which encourages the student to transition smoothly across the illusion boundary. Conversely, the edge subset and the guard ring regularize the student to match the teacher's second-order structure, preserving structural boundaries and suppressing halo-like artifacts.

\paragraph{Anchoring the re-edited region to local geometric reality.}
Finally, these regions collectively establish a geometry-aware transition framework that supports per-instance geometric reasoning inside the ROI. For each individual ROI polygon, the code uses its per-instance ring neighborhood and teacher-normalized depth to fit a small set of local plane hypotheses, which then inform the gating-based re-editing loss. Ultimately, this structured approach suppresses hallucinated geometry inside the illusion ROI while maintaining stable, teacher-guided behavior in the surrounding scene and across the ROI boundary.

\begin{table*}[t]
\centering
\caption{\textbf{Impact of extended training duration (Best Checkpoint).} While training for more epochs marginally improves hallucination metrics (DCS/CCS $\downarrow$), it degrades background knowledge preservation ($R^2 \uparrow$) and yields mixed results in DA-2k and \textit{Depth-In-the-Wild}. Epoch 1 represents the optimal trade-off. For the downstream datasets, NYUv2 and DA-2k report pairwise accuracy (acc), while DIW reports the Weighted Human Disagreement Rate (WHDR). \textit{Note: Epoch 3 reused the best checkpoint from Epoch 2 because no better checkpoint was found}}
\vspace{-2mm}
\setlength{\tabcolsep}{6pt}
\resizebox{\textwidth}{!}{%
\begin{tabular}{c c c c c c c c c c c}
\toprule
Epoch &
$d_{\textbf{cluster}}$$\downarrow$ & $d_{\textbf{avg}}$$\downarrow$ & \textbf{DCS}$\downarrow$ &
$D_{\textbf{cluster}}$$\downarrow$ & $D_{\textbf{avg}}$$\downarrow$ & \textbf{CCS}$\downarrow$ &
$R^2$ [\%] &
NYUv2 [\%] &
DA-2k [\%] &
DIW [\%]$\downarrow$ \\
\midrule
\textbf{1} &
28.55 & 30.08 & 58.64 &
$9.174\!\times\!10^{-5}$ &
$9.894\!\times\!10^{-5}$ &
$1.907\!\times\!10^{-4}$ &
\textbf{93.95} &
89.50 &
\textbf{96.08} &
\textbf{11.475} \\

2 &
28.27 & 29.86 & 58.13 &
$9.145\!\times\!10^{-5}$ &
$9.914\!\times\!10^{-5}$ &
$1.906\!\times\!10^{-4}$ &
93.59 &
\textbf{89.76} &
\textbf{96.08} &
11.649 \\

3* &
28.27 & 29.86 & 58.13 &
$9.145\!\times\!10^{-5}$ &
$9.914\!\times\!10^{-5}$ &
$1.906\!\times\!10^{-4}$ &
93.59 &
\textbf{89.76} &
\textbf{96.08} &
11.649 \\

4 &
\textbf{26.08} & \textbf{27.59} & \textbf{53.66} &
$\mathbf{8.083\!\times\!10^{-5}}$ &
$\mathbf{8.860\!\times\!10^{-5}}$ &
$\mathbf{1.694\!\times\!10^{-4}}$ &
93.49 &
89.42 &
95.70 &
11.860 \\
\bottomrule
\end{tabular}}
\label{tab:trainmore}
\end{table*}

\section{Second-Order Metrics Test}
\label{subsec:dbg-metrics}

\begin{wrapfigure}{r}{0.65\textwidth}
\vspace{0mm}
    \centering
    \includegraphics[width=0.98\linewidth]{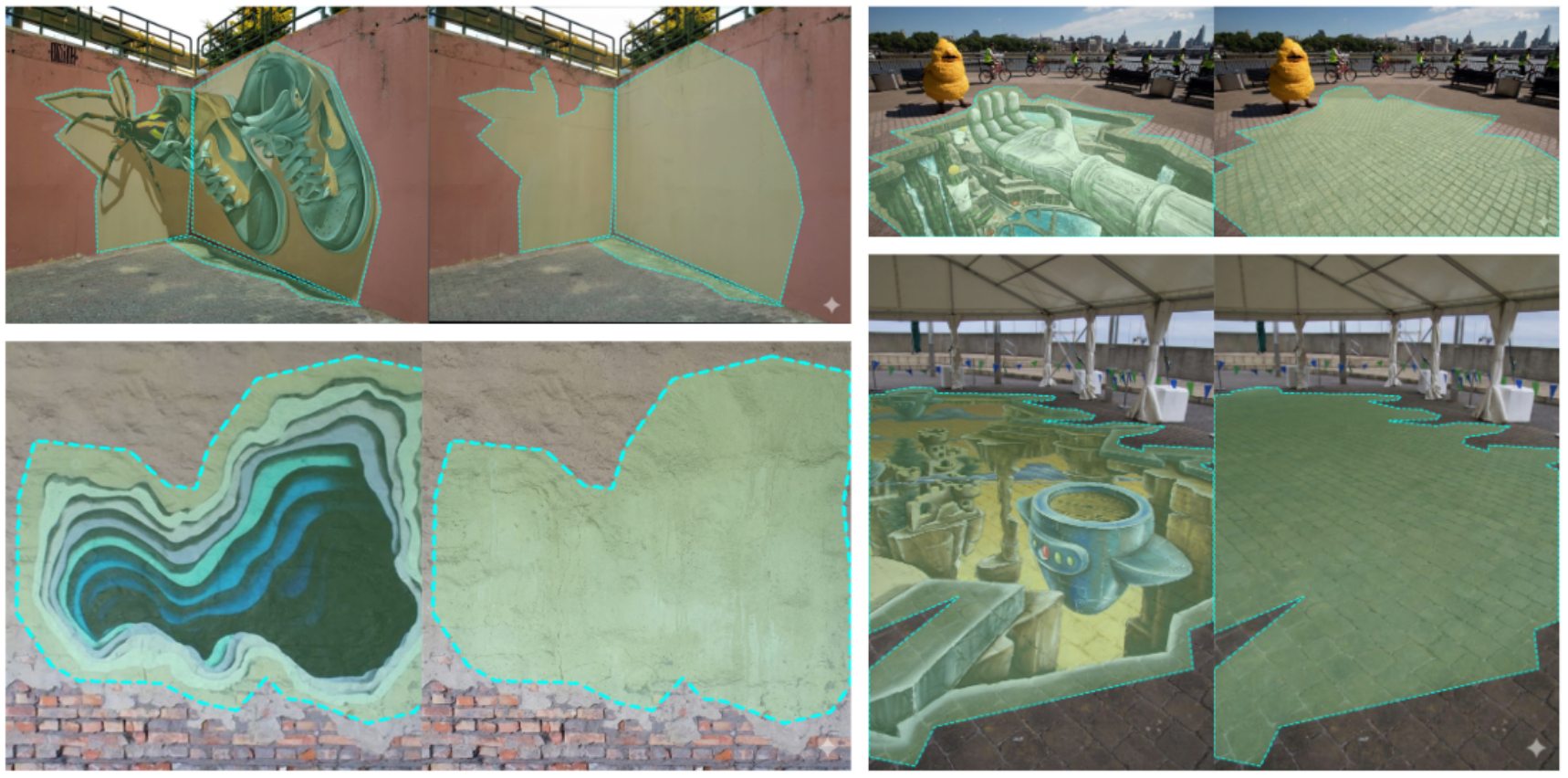}\vspace{-2mm}
\caption{\textbf{Scenes with illusory regions and edited counterparts without illusion}. Image editing was used to remove the illusory content and synthesize a plausible non-illusory texture while keeping the surrounding scene as consistent as possible.}
    \label{fig:metrics_noise_test}
    \vspace{-6mm}
\end{wrapfigure}

\paragraph{Isolating the illusion response via controlled scene manipulation.}
To validate that our proposed metrics isolate structural hallucinations rather than generic image artifacts, we designed a targeted control experiment. By comparing scenes with and without their deceptive textures, we can better isolate the metrics' response to the illusion itself from baseline network behavior. Specifically, we evaluated four scenes (Fig.~\ref{fig:metrics_noise_test}) for which we had both the original illusion image and a corresponding non-illusory counterpart. The non-illusory images were generated by removing the deceptive content while keeping the surrounding scene context as consistent as possible (via Gemini Nano Banana). Each non-illusory image was spatially aligned to its original counterpart so that both were evaluated on the same spatial support.

\paragraph{Evaluating metrics against image degradation baselines.}
With the spatial support standardized, we evaluated our deviation and confusion metrics across different image conditions to reduce trivial confounding factors. This involved testing not only the clean non-illusory images but also synthetically degraded variants. We computed DCS and CCS from paired full-image and crop predictions, using the annotated ROI union intersected with the valid overlap region as the evaluation mask. Furthermore, we evaluated Gaussian-perturbed versions of the non-illusory images. This additional control tested whether our metrics were merely sensitive to mild image degradation or high-frequency perturbations, rather than to illusion-induced geometric inconsistency.

\begin{wrapfigure}{r}{0.65\textwidth}
\vspace{-8mm}
    \centering
    \includegraphics[width=0.98\linewidth]{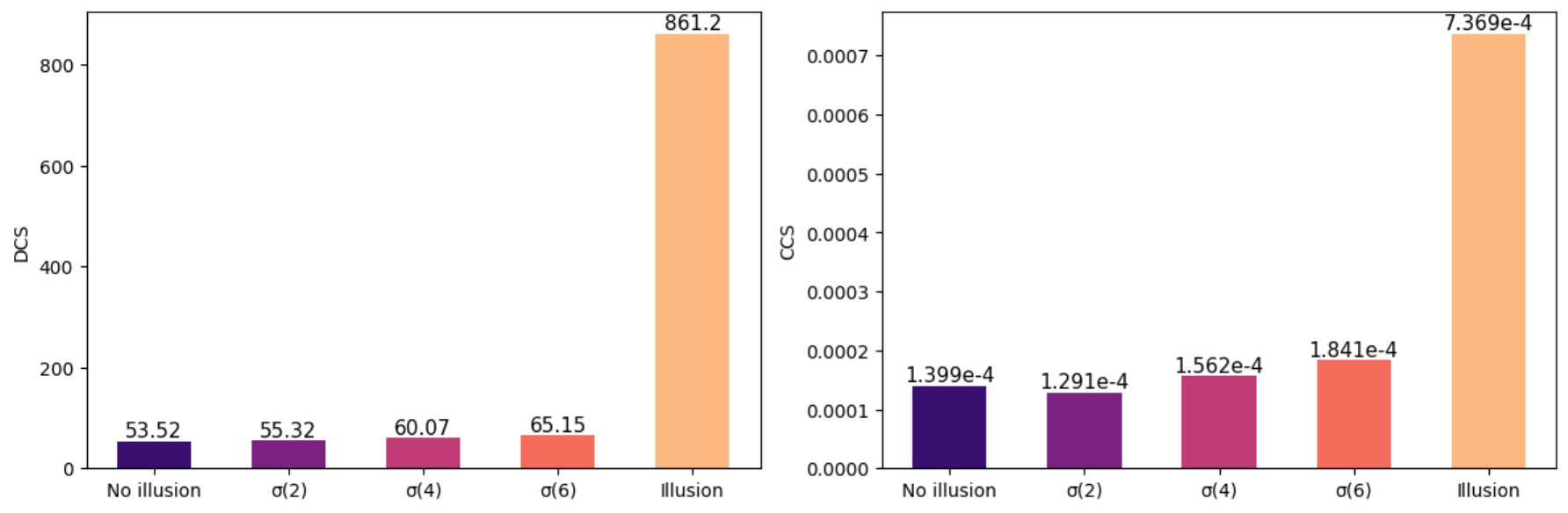}\vspace{-2mm}
\caption{\textbf{Control experiments across 4 scenes}. DCS and CCS are shown for the original images, non-illusory counterparts, and Gaussian-perturbed versions. While mild noise causes only small score changes, the illusion images produce substantially larger responses, supporting that proposed metrics are driven primarily by illusion-induced depth inconsistency}
    \label{fig:metrics_noise_chart}
    \vspace{-6mm}
\end{wrapfigure}
\paragraph{Validating metric robustness and novelty.}
The quantitative results show a much larger metric response for the illusory images than for the matched non-illusory controls or their noisy variants. Aggregated over $4$ scenes and $16$ crops (Fig.~\ref{fig:metrics_noise_chart}), the matched no-illusion control remained low ($\mathrm{DCS}=53.52$, $\mathrm{CCS}=1.40\times10^{-4}$). Introducing Gaussian noise caused only marginal score increases (reaching at most $\mathrm{DCS}=65.15$ and $\mathrm{CCS}=1.84\times10^{-4}$ at $\sigma=6$). In contrast, the illusion images triggered a large spike in both metrics ($\mathrm{DCS}=861.24$, $\mathrm{CCS}=7.37\times10^{-4}$), yielding aggregate illusion--control gaps of $\Delta\mathrm{DCS}=807.72$ and $\Delta\mathrm{CCS}=5.97\times10^{-4}$. 

These findings suggest that, under matched scene context and identical crop support, DCS and CCS are relatively insensitive to generic alignment artifacts, crop geometry, and mild pixel-level noise in this control setting. Overall, this control provides evidence that our proposed metrics function as targeted probes of illusion-induced depth inconsistencies, which standard pixel-wise MDE evaluations are not designed to capture directly.

\section{Visualization of Second-Order Metrics Across Models}
\label{subsec:vis-metrics}

\begin{figure}[t]
    \centering
    \includegraphics[width=\linewidth]{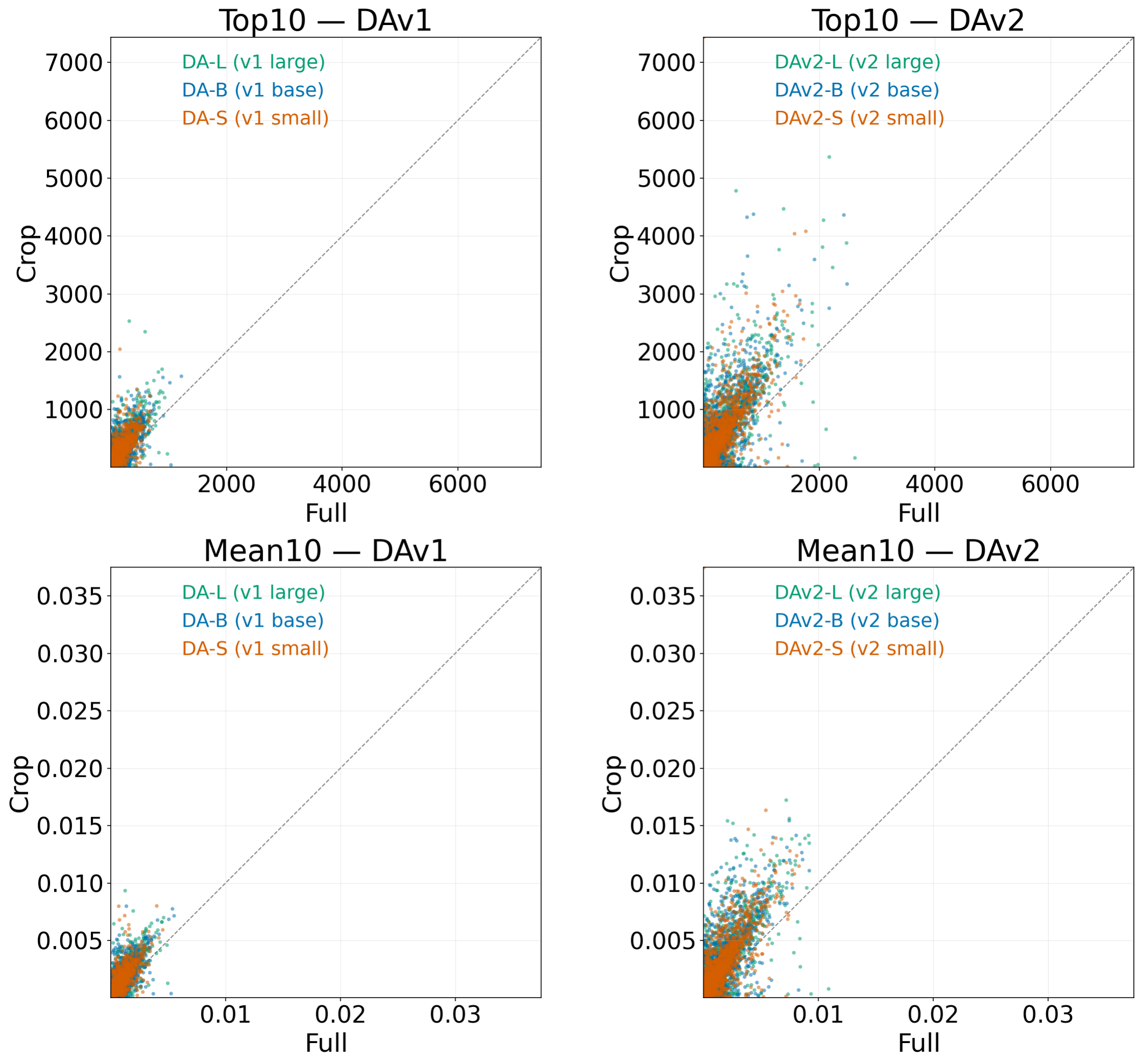}
    \caption{\textbf{Second-order Response Analysis: Relative Models.} We plot the projected Second-order responses for full-context (x-axis) vs. crop-context (y-axis) inputs. The first column shows Depth Anything (v1), and the second shows DAv2. Colors denote model size: Small (orange), Base (blue), Large (green). The systematic upward shift above the diagonal demonstrates that hallucinations intensify when context is removed.}
    \label{fig:lap_rel_m_vis}
\end{figure}

\begin{figure}[t]
    \centering
    \includegraphics[width=\linewidth]{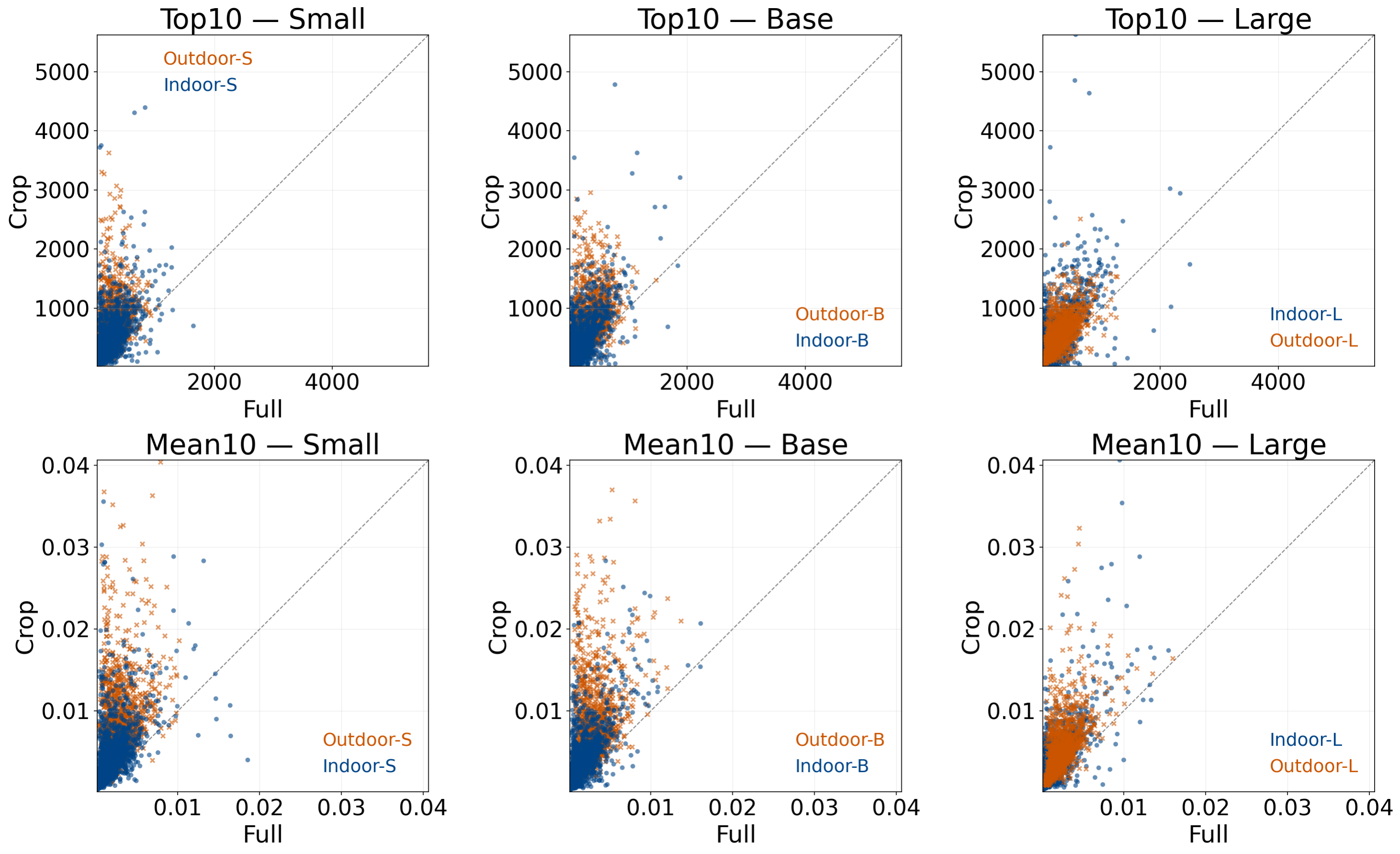}
    \caption{\textbf{Second-order Response Analysis: Metric Models.} A comparison of Indoor (orange) and Outdoor (blue) DAv2 variants across sizes (S/B/L). Indoor models exhibit tighter clustering near the origin, indicating lower hallucination intensity. Outdoor models show high dispersion at smaller sizes, contracting only at the Large scale due to structural collapse (outputting flat artifacts). Points are plotted with denser clusters on top to maximize visibility.}
    \label{fig:lap_met_m_vis}
\end{figure}

To visualize the structural behavior of hallucinations, we plot the aligned full-view versus crop-view second-order responses in the $(\text{full}, \text{crop})$ plane for both \emph{relative} models (DA/DAv2) and \emph{metric} DAv2 models (Fig.~\ref{fig:lap_rel_m_vis} and Fig.~\ref{fig:lap_met_m_vis}). Each data point represents a single benchmark instance (a full image and its paired crop), with responses computed on the union ROI mask for that instance. The coordinates correspond to the projected $\mathrm{top}_{\scriptscriptstyle 10}$ (sum of top 10\% magnitudes) or $\mathrm{mean}_{\scriptscriptstyle 10}$ (mean of the top 90\%) Second-order response within the union ROI mask, computed after ROI-conditional quantile normalization.

A key observation across all variants is that the point clouds lie systematically above the diagonal ($y=x$). This indicates that Second-order energy - and thus geometric hallucination - is consistently \emph{stronger} under reduced context (crop) than under full context. This confirms the context-dependent nature of the 3D Mirage failure mode.

\vspace{0.5em}
\noindent\textbf{Relative Models.}
As shown in Fig.~\ref{fig:lap_rel_m_vis}, the point clouds for Depth Anything v1 (DA) are notably tighter and clustered closer to the origin compared to DAv2 across all model sizes. This suggests that DAv2 models are more susceptible to strong hallucinations than their predecessors. Furthermore, all variants exhibit a distinct upward skew, confirming that removing context exacerbates the prediction of spurious non-planar geometry.

\begin{figure}[t]
    \centering
    \includegraphics[width=0.95\linewidth]{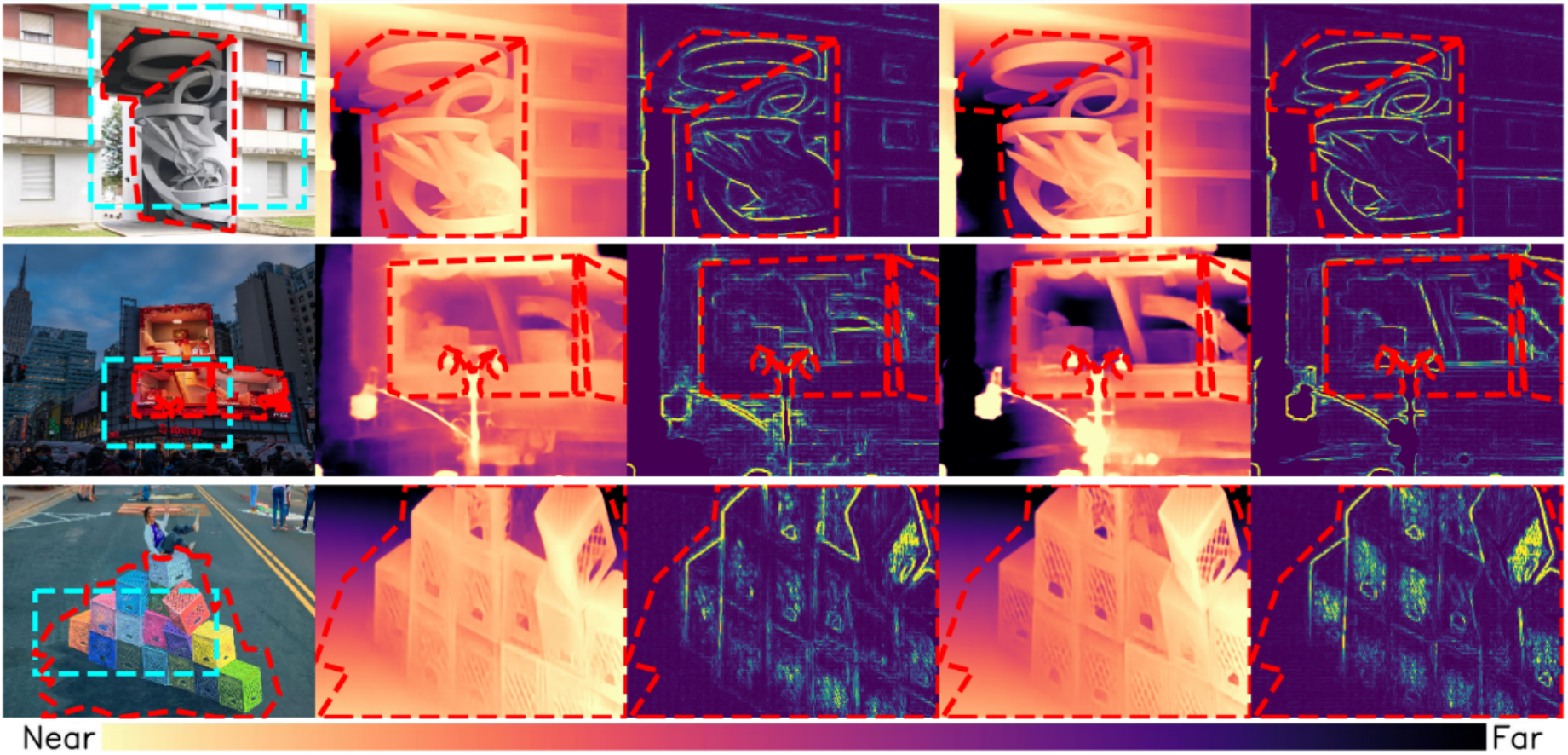}
    \caption{\textbf{Impact of Model Capacity on Edge Fidelity.} From left to right: Original RGB, Depth output and 2nd-order response of DAv2-Indoor-Base(IB), then DAv2-Indoor-Large(IL). The color bar denotes relative depth. The Large model's higher fidelity resolves sharper edges, inadvertently leading to higher Second-order energy scores ($top_{10}$) in the illusion region.}
    \label{fig:ilworse}
\end{figure}

\vspace{0.5em}
\noindent\textbf{Metric Models.}
Figure~\ref{fig:lap_met_m_vis} illustrates distinct behaviors between Indoor and Outdoor training regimes.
For \textbf{Indoor} models, the Small and Base variants exhibit compact clusters near the origin. However, the transition from Base to Large results in increased dispersion for the $\mathrm{top}_{\scriptscriptstyle 10}$ metric. Qualitative analysis (Fig.~\ref{fig:ilworse}) suggests this dispersion stems from the Large model's higher detail/edge fidelity, which captures sharper (albeit hallucinated) gradients.

\begin{figure}[t]
    \centering
    \includegraphics[width=0.95\linewidth]{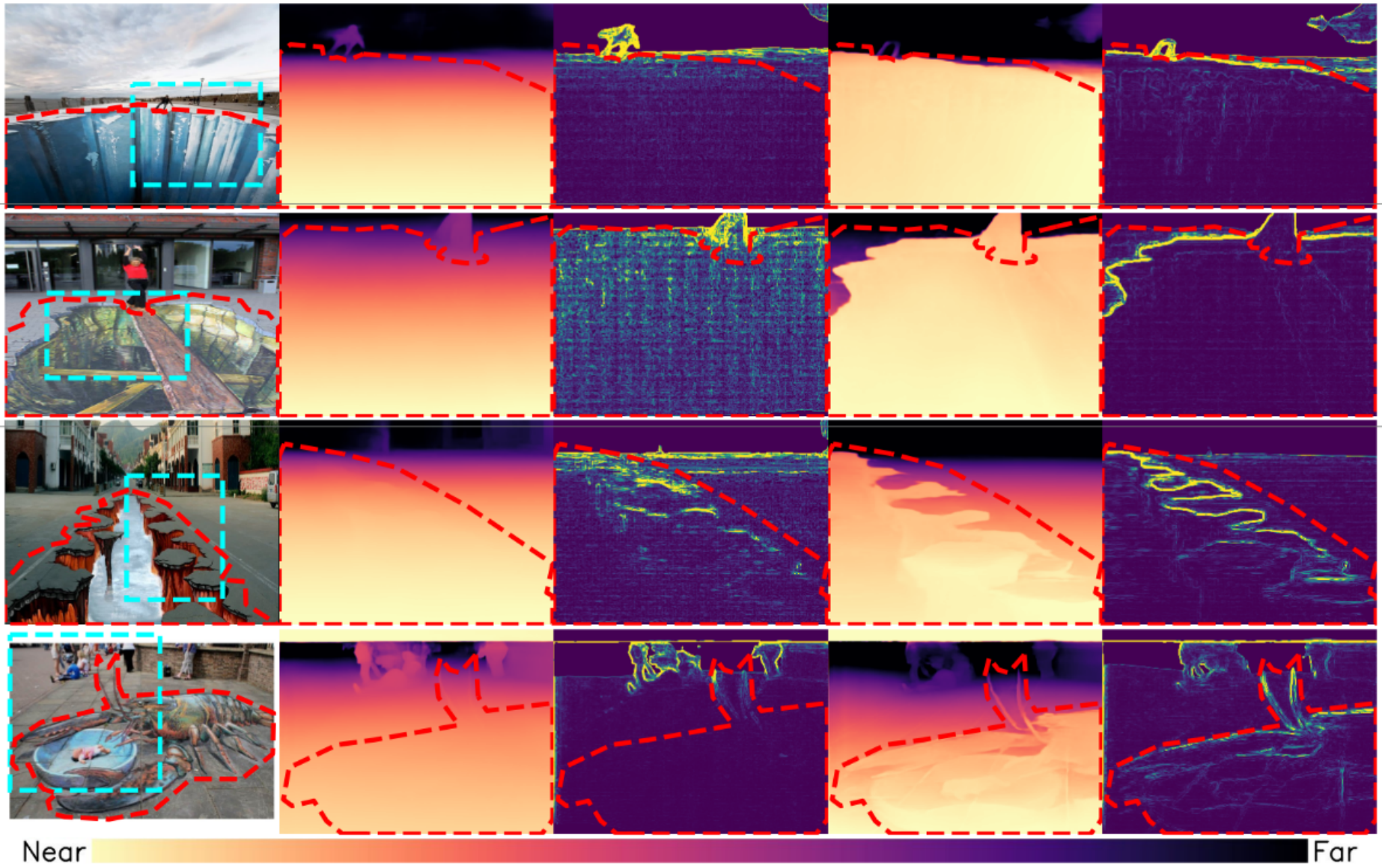}
    \caption{\textbf{Domain Specialization Comparison.} From left to right: Original RGB, depth-output and corresponding 2nd-order response of DAv2-Indoor-Base(IB), then DAv2-Outdoor-Base(OB) on the same input. The Indoor variant successfully predicts a planar surface, whereas the Outdoor variant strongly hallucinates a depression.}
    \label{fig:ibw_obf}
\end{figure}

\begin{figure}[t]
    \centering
    \includegraphics[width=0.95\linewidth]{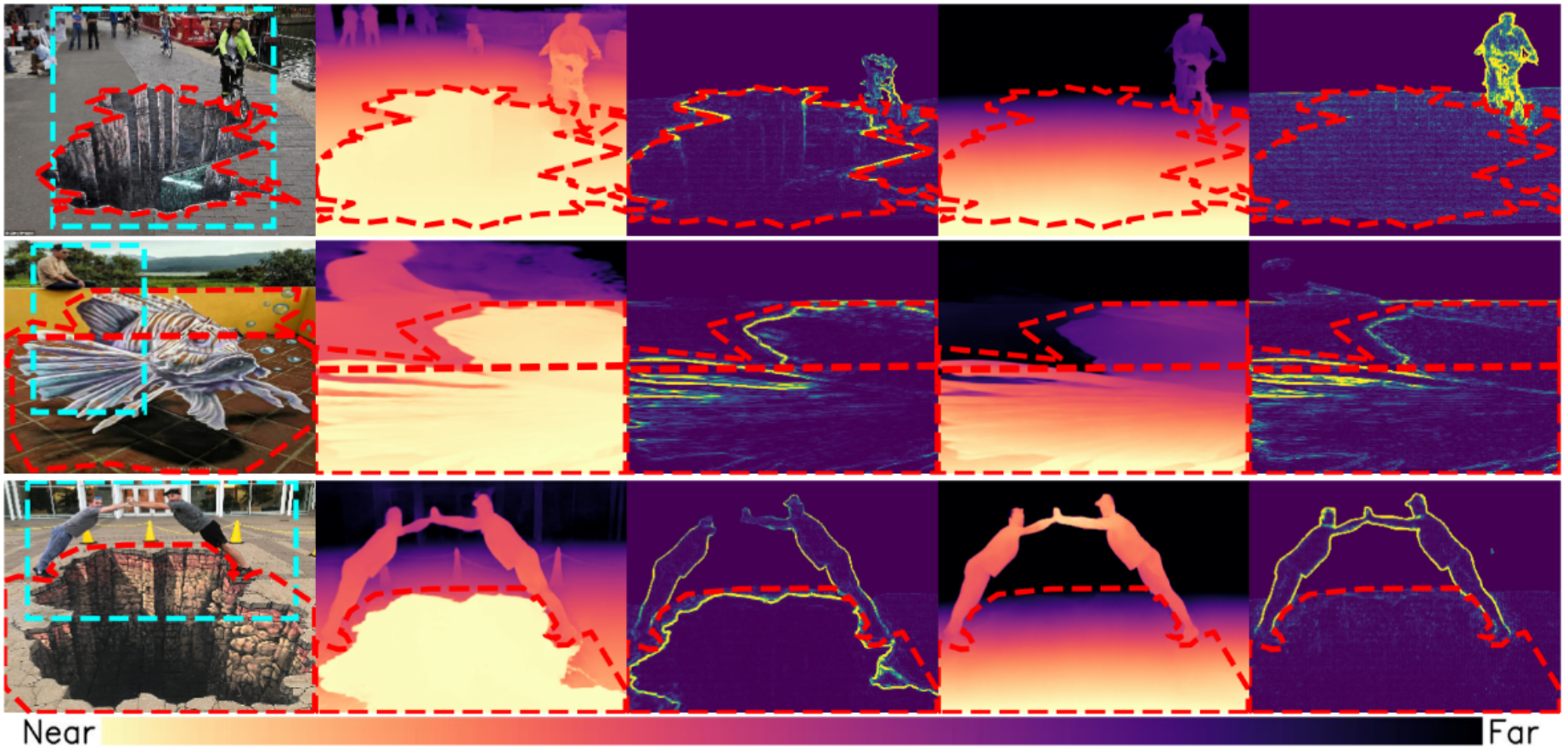}
    \caption{\textbf{Failure Mode: The ``Vertical Obstacle'' Bias.} From left to right: Original RGB, depth-output and corresponding 2nd-order response of DAv2-OB, then DAv2-IB. DAv2-OB hallucinates a low-variance, near-vertical patch, effectively treating the illusion as an immediate obstacle. The DAv2-IB model modestly hallucinates but still preserves general geometry.}
    \label{fig:obfailin}
\end{figure}

\begin{figure}[t]
    \centering
    \includegraphics[width=0.95\linewidth]{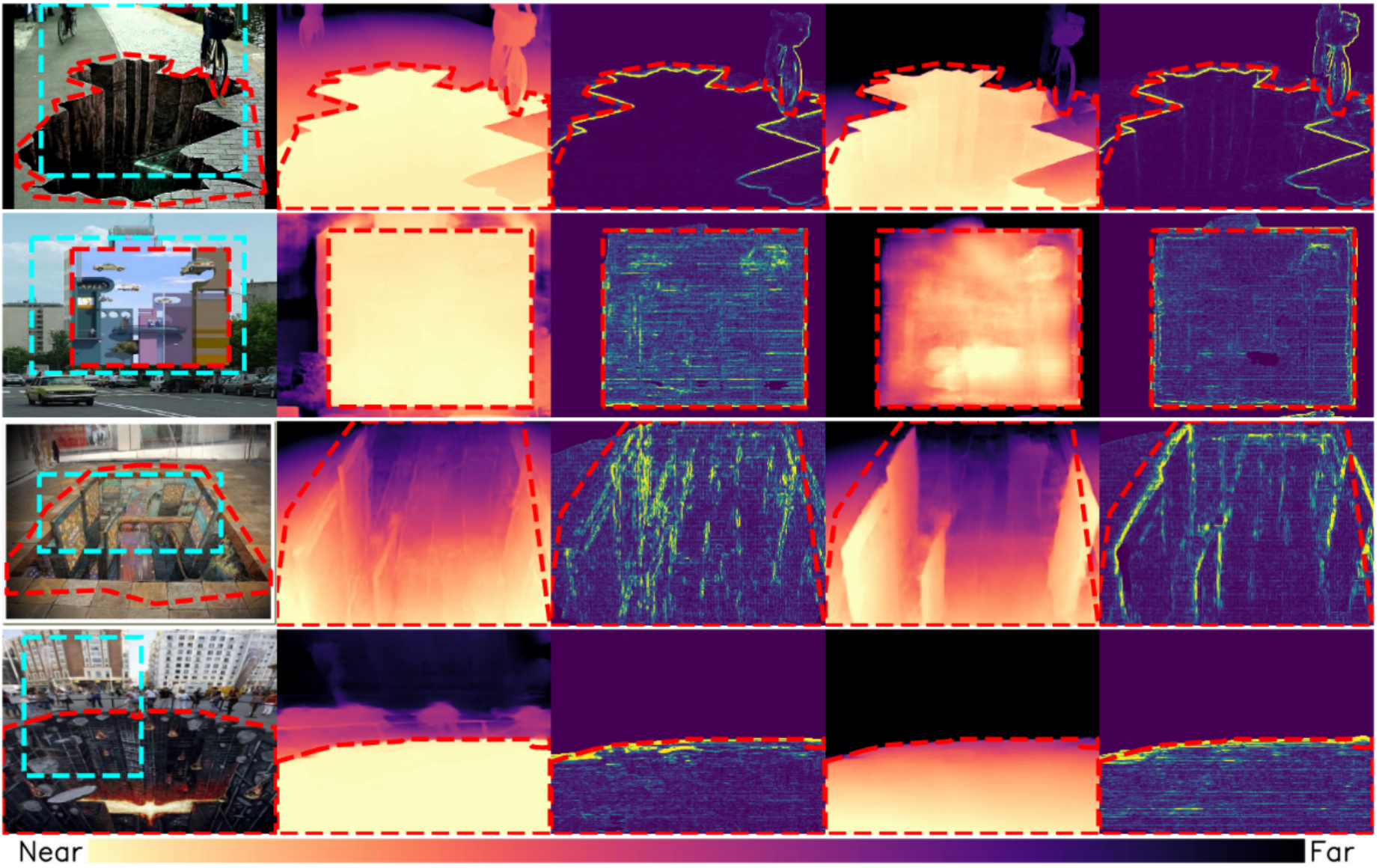}
    \caption{\textbf{Failure Mode: Structural Collapse in Large Outdoor Models.} From left to right: Original RGB, depth-output and corresponding 2nd-order response of DAv2-OL, then DAv2-OB. Top rows: Full context; Bottom rows: Crop context. In high-ambiguity scenarios, the Large Outdoor model (OL) abandons geometric plausibility, filling the region with a monotonic, texture-less patch that ignores both real geometry and the illusion.}
    \label{fig:outbigfail}
\end{figure}

\begin{figure}[t]
    \centering
    \includegraphics[width=0.95\linewidth]{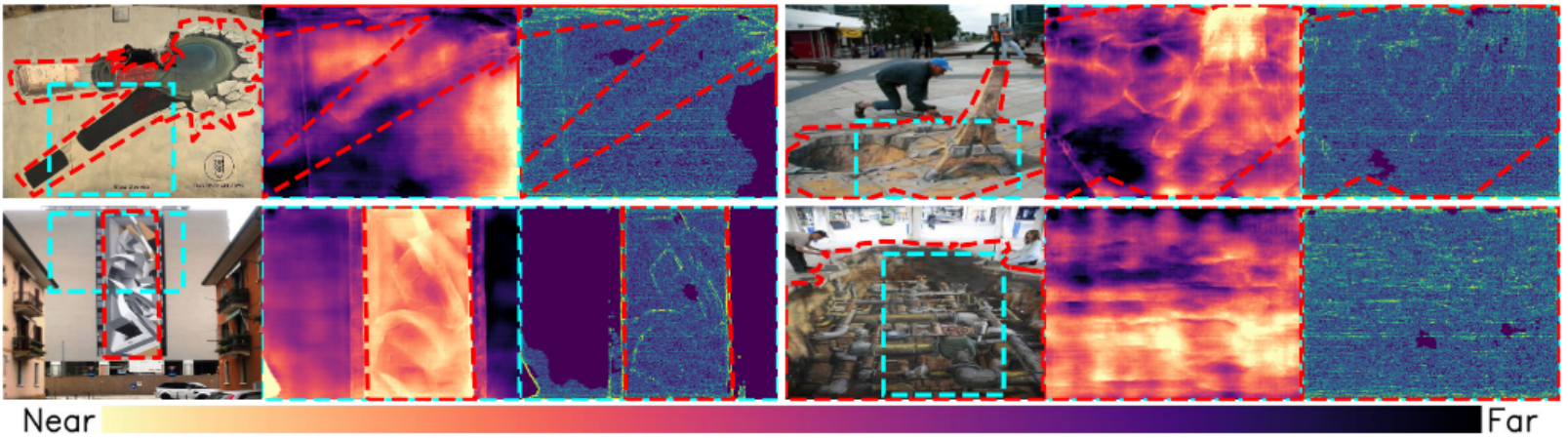}
    \caption{\textbf{Failure Mode: Noisy Collapse.} From left to right: Original RGB, depth-output and corresponding 2nd-order response of DAv2-OB. When a scene with reduced context does not conform to the standard outdoor layouts (e.g., distinct sky/ground separation), the Outdoor-Base model fails to converge, outputting incoherent, noisy depth clouds.}
    \label{fig:outmess}
\end{figure}

\textbf{Outdoor} models, conversely, show clusters that are initially dispersed but contract toward the origin as capacity increases (Base $\to$ Large). Our analysis reveals two distinct failure modes driving this behavior. First, the Outdoor-Base (OB) model frequently ``fills in'' the illusion region with a constant-depth patch, effectively treating the illusion as a vertical obstacle (Fig.~\ref{fig:obfailin}). Conversely, when the OB model successfully ignores the illusion (Fig.~\ref{fig:ibw_obf}), it often relies on specific side-context cues (e.g., horizons, curbs). When scenes deviate from these deterministic layouts - or when context is sufficiently reduced - the Outdoor models tend to suffer from \emph{structural collapse}, outputting noisy, incoherent depth clouds (Fig.~\ref{fig:outmess}).
By the Large size (OL), the Outdoor point clouds tighten, resembling the Indoor distribution. Qualitative evidence (Fig.~\ref{fig:outbigfail}) suggests this is because the OL model resolves hallucinations with high confidence, replacing the illusion with smooth, monotonic patches that ignore both real geometry and local context cues.

\begin{figure*}[t]
    \centering
    \includegraphics[width=0.96\linewidth]{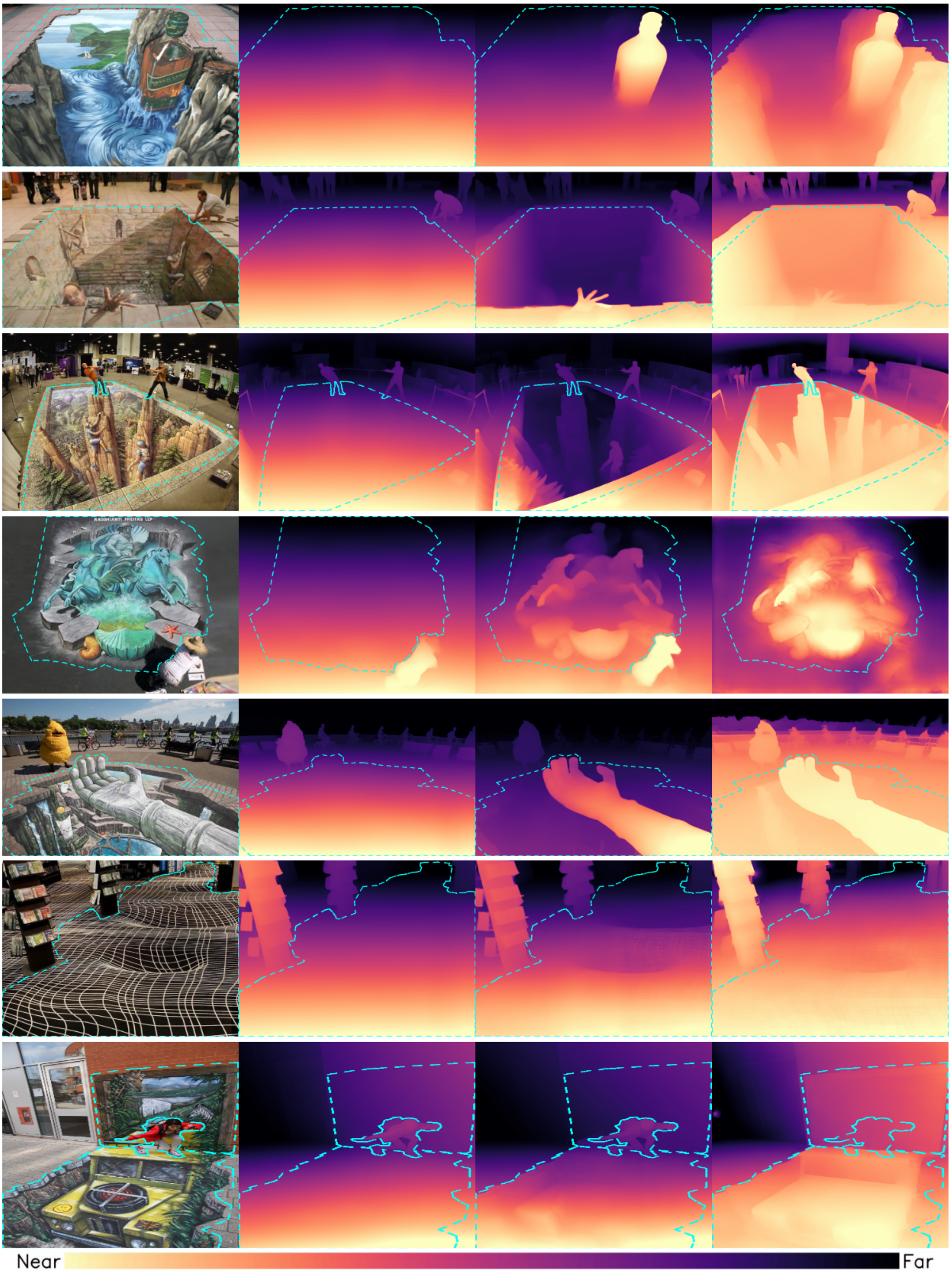}\vspace{-2mm}
    \caption{\textbf{Performance Comparison 1.} From left to right: RGB with ROI marking, then depth output of: Ours, DepthPro, DepthFM. Visual comparison of our Grounded Self-Distillation method against DepthPro and DepthFM on the test set. Our method maintains 3D structural integrity while baselines exhibit hallucination.}
    \label{fig:vissup1}
\end{figure*}

\begin{figure*}[t]
    \centering
    \includegraphics[width=0.95\linewidth]{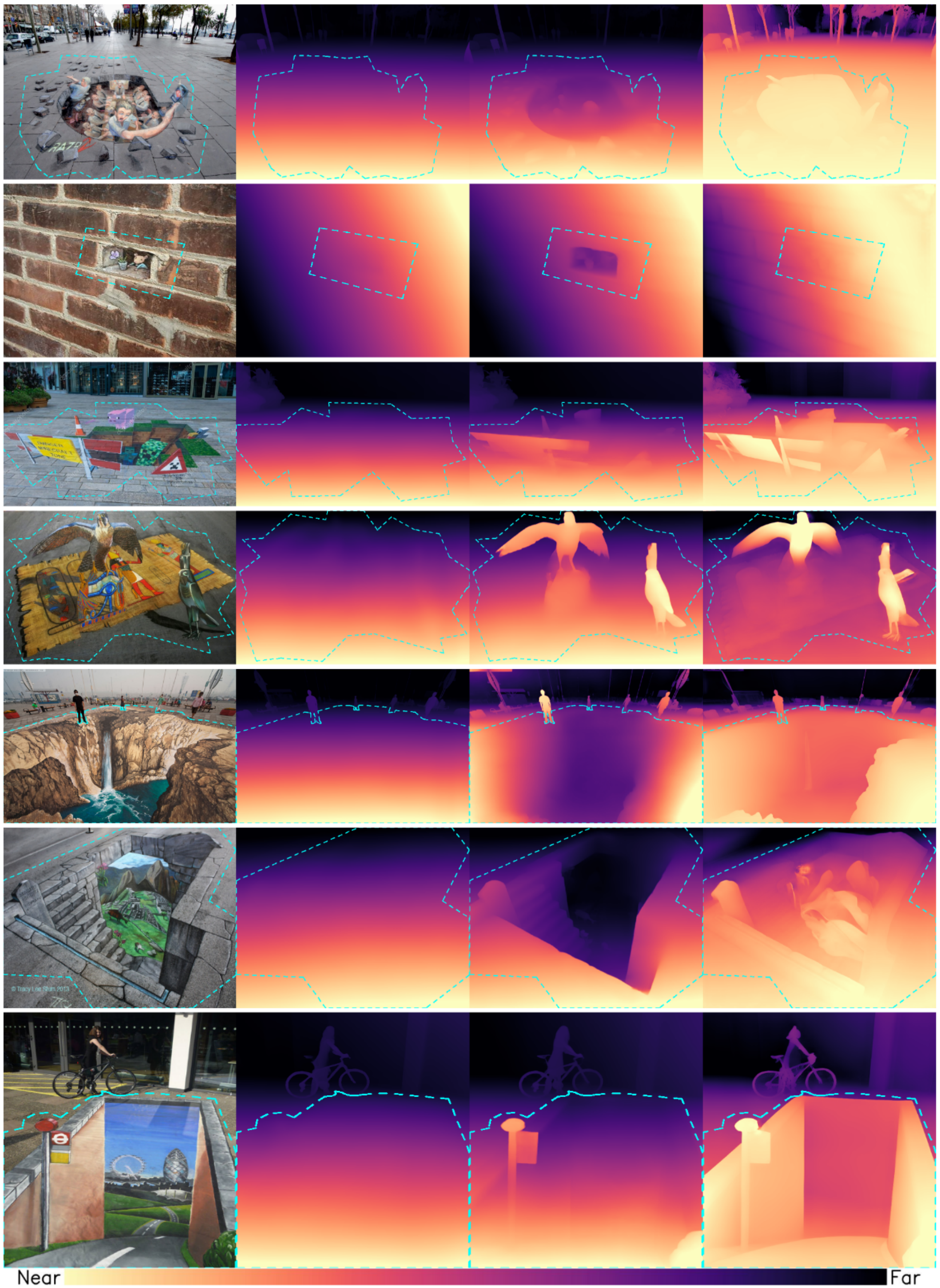}\vspace{-2mm}
    \caption{\textbf{Performance Comparison 2.} Further visual examples comparing our method against DepthPro and DepthFM on the test set, highlighting robustness in challenging illusion scenarios.}
    \label{fig:vissup2}
\end{figure*}

\begin{figure}[t]
    \centering
    \includegraphics[width=0.95\linewidth]{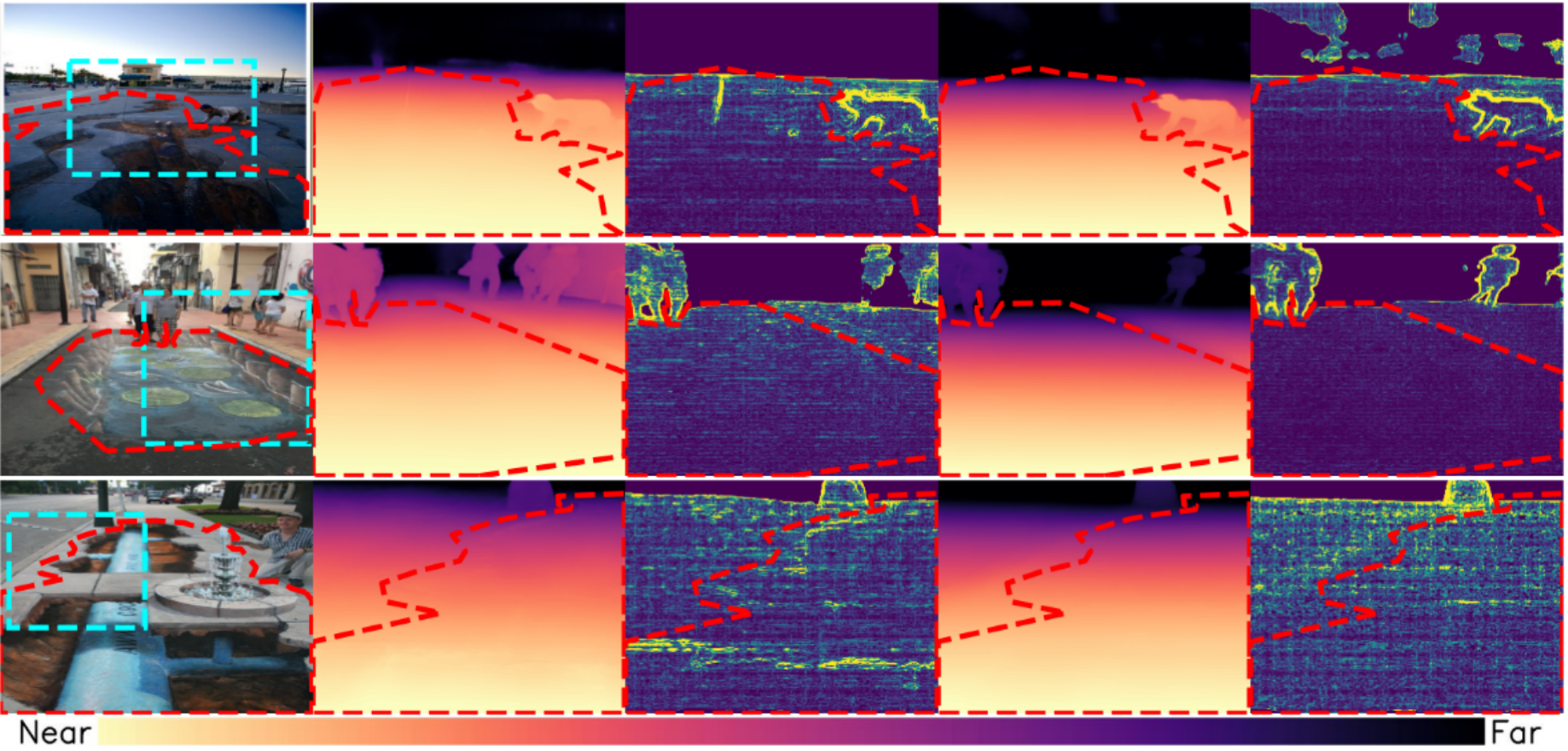}
    \caption{\textbf{Contextual Disambiguation in Outdoor Models.} From left to right: Original RGB, depth-output and corresponding 2nd-order response of DAv2-OB, then DAv2-IB. The Outdoor model occasionally ignores the illusion only when strong side-context cues (e.g., sidewalks, road horizons) are present within the crop, but overall tends to return nearer values than IB variant.}
    \label{fig:obworkin}
\end{figure}

\section{DCS: Hallucination Magnitude}
Table 1 in the main text quantifies hallucination magnitude via the Deviation Composite Score (DCS). Here we analyze the underlying drivers of these scores.

\begin{figure}[t]
    \centering
    \includegraphics[width=0.95\linewidth]{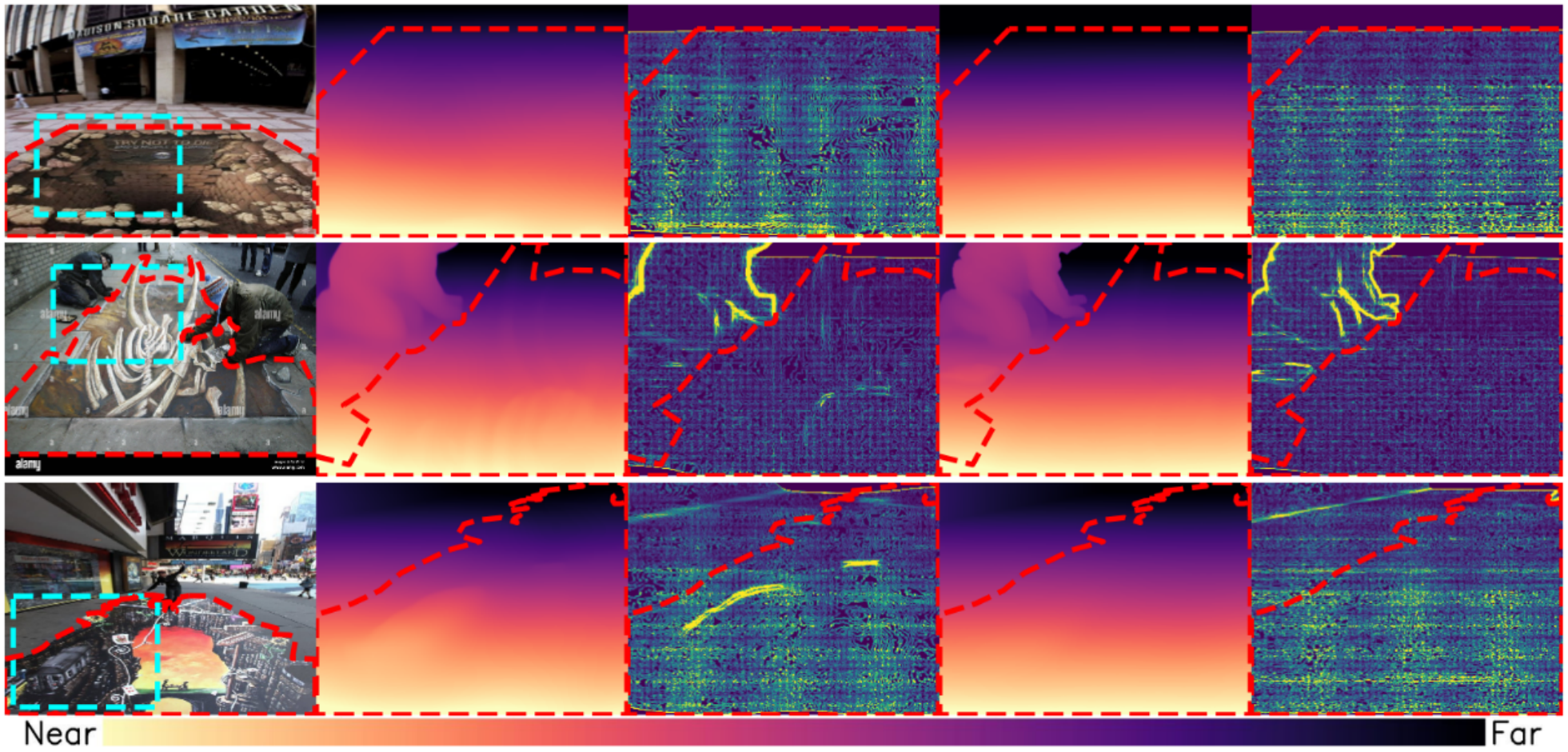}
    \caption{\textbf{Qualitative Comparison: Low Hallucination Regime.} From left to right: Original RGB, depth-output and corresponding 2nd-order response of DA-B, then DAv2-B. Both models perform well, with improvement in DAv2. Note the high similarity in both depth and Second-order outputs.}
    \label{fig:rel_low}
\end{figure}

\begin{figure*}[t]
    \centering
    \includegraphics[width=\linewidth]{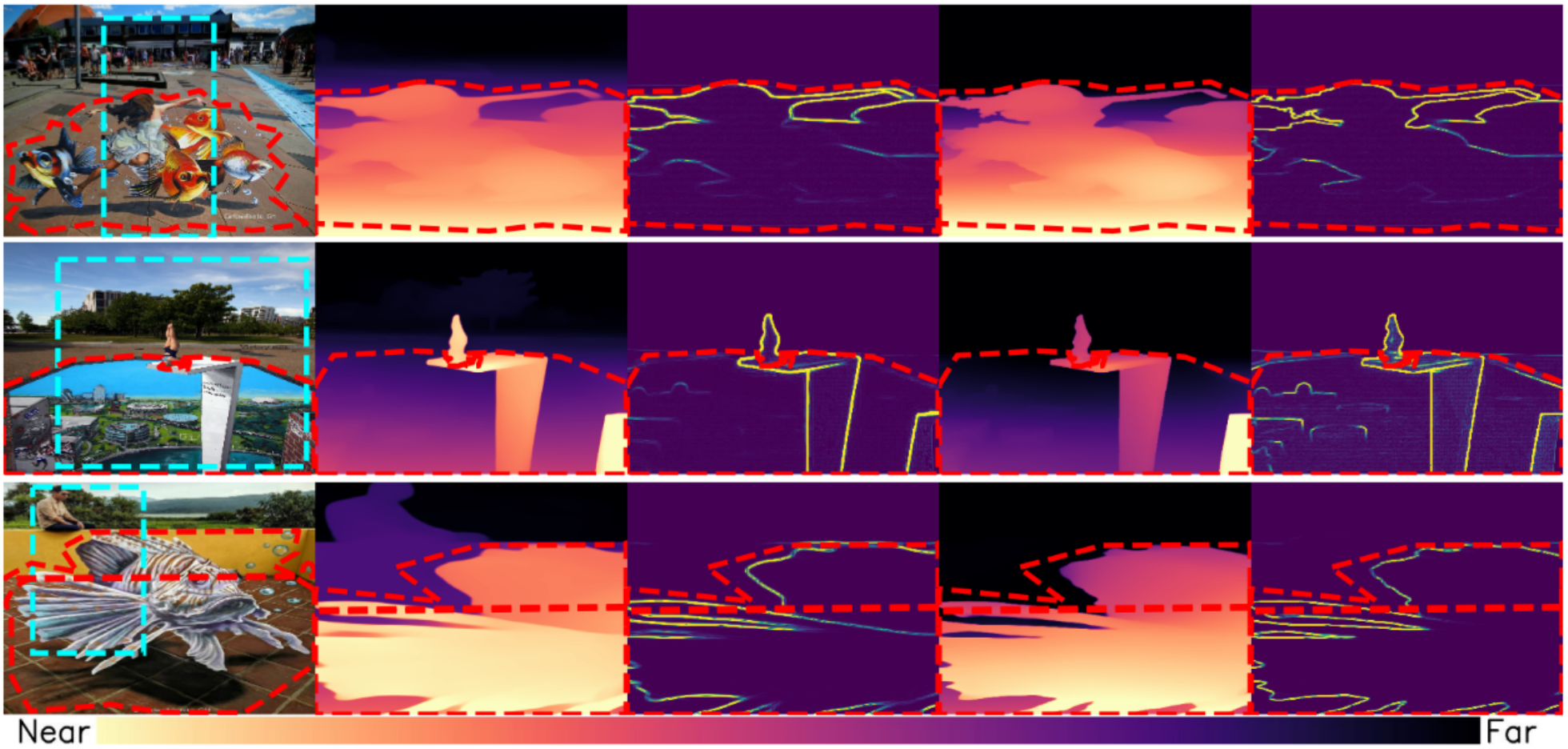}
    \caption{\textbf{Qualitative Comparison: High Hallucination Regime.} From left to right: Original RGB, depth-output and corresponding 2nd-order response of DA-Base, then DAv2-B. When both models hallucinate, DAv2-Base generates significantly sharper geometric details and edges, resulting in a higher overall DCS.}
    \label{fig:rel_high}
\end{figure*}

\noindent\textbf{Relative Models.}
DA(v1) achieves markedly lower DCS than DAv2 across all sizes. We investigated whether this gap stems from DAv2's synthetic teacher bias or simply higher output fidelity. Qualitative comparisons (Fig.~\ref{fig:rel_low} and \ref{fig:rel_high}) reveal that while DA-Base and DAv2-Base perform similarly on low-hallucination samples, DAv2-Base generates significantly sharper, higher-fidelity hallucinations on difficult samples. This suggests that the higher DCS in v2 models is driven by their improved capability to resolve (spurious) high-frequency details, rather than solely by a shift in training distribution.

\noindent\textbf{Metric Models.}
Indoor models consistently achieve lower DCS than Outdoor models (e.g., 39\% lower for Small, 51\% lower for Base). This performance gap is likely attributable to the Indoor training data, which contains semantically diverse, textured, and cluttered scenes. This diversity forces the model to learn robust local geometric cues.
\begin{itemize}
    \item \textbf{Indoor Scaling:} Performance peaks at the \textbf{Base} size. Large models exhibit slightly higher DCS due to their tendency to resolve hallucinations with sharper edges.
    \item \textbf{Outdoor Scaling:} Performance improves primarily at the \textbf{Large} scale (OL reduces DCS by $\sim$25\% vs. OS/OB). However, this numerical improvement often masks a qualitative degradation: In high confusion cases, OL models tend to collapse into ``safe,'' low-variance depth patches that lack geometric detail, rather than correctly recovering the planar surface.
\end{itemize}

\section{CCS: Context Dependence and Stability}
\noindent\textbf{Relative Models.}
DA(v1) models exhibit significantly greater stability, with CCS values $\sim$55--60\% lower than DAv2, likely resulting from the lower overall fidelity of the older variant's hallucinations. In relative models, the $D_{\text{avg}}$ component consistently exceeds $D_{\text{cluster}}$ ($\approx$10--30\%), implying that context instability manifests as dispersed, per-pixel variance rather than a systematic shift of the entire depth distribution.

\begin{figure}[t]
    \centering
    \includegraphics[width=0.95\linewidth]{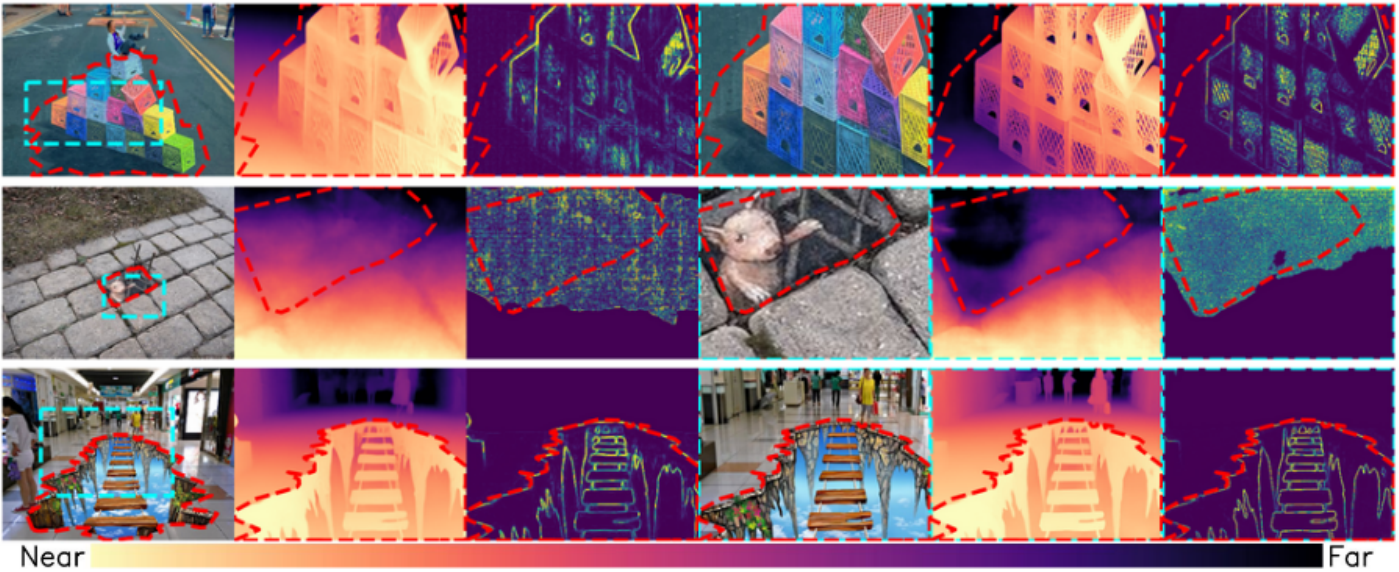}
    \caption{\textbf{Indoor Stability.} From left to right: Original RGB, depth-output and corresponding 2nd-order response of DAv2-IB. Scenes where DAv2-IB exhibits high context dependence (large diagonal shift in Second-order space). Even so, the structural form of the hallucination remains relatively consistent.}
    \label{fig:hi_ccs_indoor}
\end{figure}

\begin{figure}[t]
    \centering
    \includegraphics[width=0.95\linewidth]{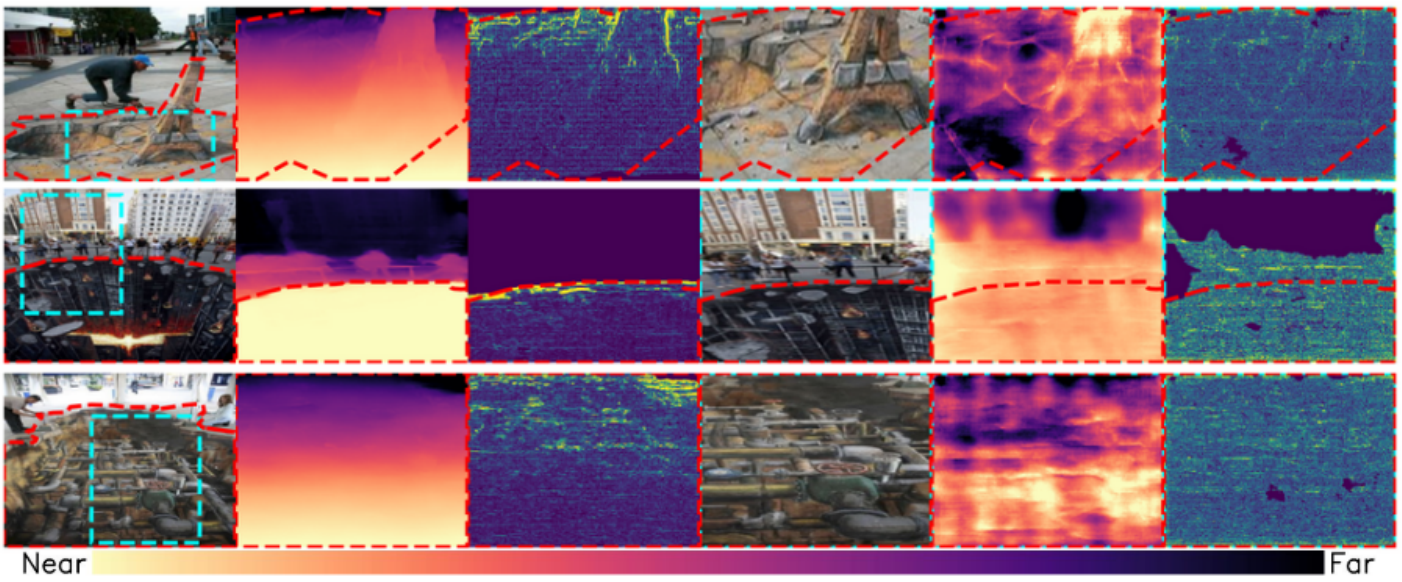}
    \caption{\textbf{Outdoor Instability.} From left to right: Original RGB, depth-output and corresponding 2nd-order response of DAv2-OB. Scenes where DAv2-OB exhibits extreme context dependence. Context removal causes a shift from structured prediction to random artifacts.}
    \label{fig:hi_ccs_outdoor}
\end{figure}

\begin{figure*}[t]
    \centering
    \includegraphics[width=\linewidth]{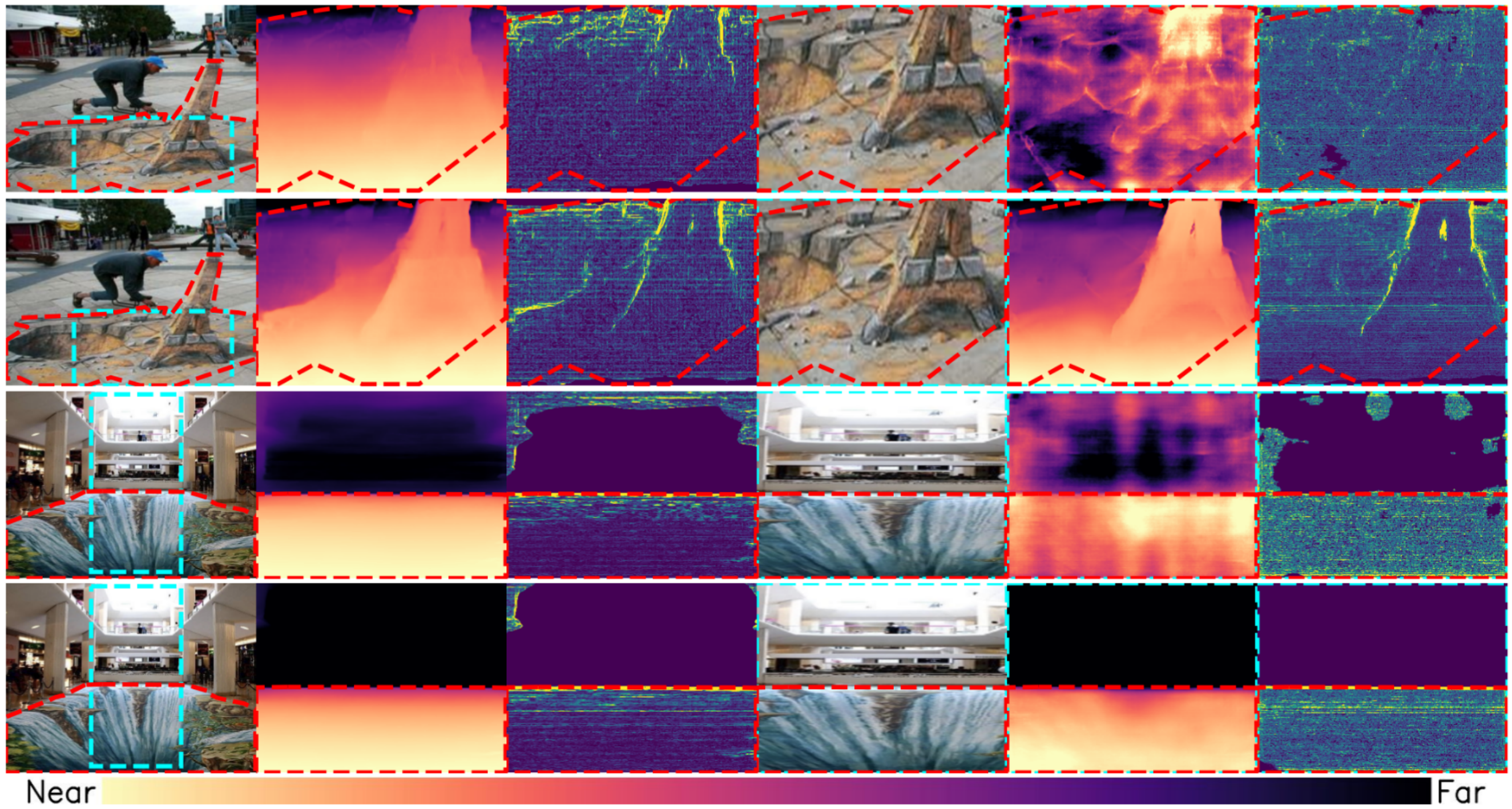}
    \caption{\textbf{Size-based Stability Improvement} From left to right: Original RGB, model depth-output and corresponding 2nd-order response given full scene, Cropped RGB, model depth-output and corresponding 2nd-order response when given crop scene. From top down: DAv2-OB(Rows 1, 3), DAv2-OL(Rows 2, 4). Scenes where DAv2-OB exhibits high context dependence (large diagonal shift in Second-order space). OL is much more \textbf{\textit{stable}} than OB, with depth outputs consistent between crop and full scene.}
    \label{fig:bl_css_red}
\end{figure*}

\noindent\textbf{Metric Models.}
Indoor models demonstrate substantially reduced sensitivity to context removal compared to Outdoor models (e.g., $\approx$60\% lower CCS for Base variants).
\begin{itemize}
    \item \textbf{Outdoor Instability:} When scenes do not conform to learned priors (e.g., road ribbons, sky-ground stratification), Outdoor models frequently exhibit mode collapse (Fig.~\ref{fig:outmess}, \ref{fig:outbigfail}). The notable reduction in CCS and bias toward Crop seen in Fig.~\ref{fig:lap_met_m_vis} is the result of Large variant's improved stability. OL model is much less likely to resolve to noisy depth cloud when confused by context cues, as seen in Fig.~\ref{fig:bl_css_red}, in the same condition as its smaller counterpart.
    \item \textbf{Indoor Stability:} Even when Indoor models hallucinate, their hallucinations remain structurally consistent between full and cropped views (Fig.~\ref{fig:hi_ccs_indoor} vs. \ref{fig:hi_ccs_outdoor}). This indicates that while they are locally deceived, they are less reliant on global context to form a coherent prediction.
\end{itemize}

\section{Data and Failure Modes}
\noindent\textbf{Benchmark coverage and scope.} The 468-scene benchmark is \textbf{80\%} outdoor, with \textbf{16\%} spanning multiple surfaces; \textbf{40\%} of scenes depict ground or walkway illusions and \textbf{8\%} large billboards. Transparent or specular materials, road potholes and curbs, cast shadows, adverse weather, and adversarially-synthesized illusions lie outside this scope and are noted as future extensions in the main-paper Limitations.

Our analysis identifies two opposing failure regimes governing 3D hallucinations:
\begin{enumerate}
    \item \textbf{Over-Capacity (Overfitting Global Priors):} Larger models (e.g., DAv2-L) tend to over-index on global semantic priors. In some settings, they amplify hallucinated curvature, increasing DCS and often CCS. However, this behavior is not uniform: in Outdoor-Large, lower CCS may also reflect a more stable but qualitatively collapsed solution, where the model fills the ROI with a smooth, low-variance patch rather than recovering the true planar structure.
    \item \textbf{Under-Capacity (Systematic Bias):} Smaller models, particularly those with strong dataset biases (e.g., Outdoor-Small), compress priors into simple heuristics. This often results in co-aligned, systematic biases with high error, rather than purely dispersed random failures.
\end{enumerate}

\section{More Related Work}

Unsupervised detection of anomalies in 3D data~\cite{bergmann2023anomaly,tu2024self,zavrtanik2024cheating,zavrtanik2024keep,gu2024rethinking,Li_2025_CVPR}, is essential for tasks ranging from industrial inspection to autonomous driving; however, the sparsity, noise, and high dimensionality of 3D point clouds present significant hurdles. Conventional approaches often pair local geometric descriptors with K-Nearest Neighbors~\cite{horwitz2023back}, yet these methods are susceptible to noise and frequently miss global context. Alternatives based on reconstruction, such as IMRNet~\cite{IMRNet}, are computationally intensive and prone to losing fine-grained details, while teacher-student architectures~\cite{bergmann2023anomaly} depend heavily on strict pose alignment. Furthermore, methods like AST~\cite{rudolph2023asymmetric} struggle to identify subtle deviations. Although recent multimodal~\cite{wang2023multimodal} and memory-augmented~\cite{cao2023complementary} strategies enhance feature representation, they remain largely local and lack explicit mechanisms for handling arbitrary poses. Similarly, EasyNet~\cite{chen2023easynet} is constrained by a limited receptive field that hinders the holistic understanding of shapes. Consequently, a major drawback of these techniques is their dependence on engineered, local features, resulting in brittleness to pose variations and poor generalization. While advanced self-supervised frameworks (e.g., R3D-AD~\cite{R3D-AD}),  foundation model-based models (e.g., MLLM-based~\cite{11033177}), and memory-based models (e.g., Reg3D~\cite{Real3d-AD}) improve robustness, they do so at a high computational cost. PASDF~\cite{Zheng_2025_ICCV} is a pioneering work to unify the 3D anomaly detection and repair via a unified continuous geometric representation. Unlike these predominantly geometry-focused approaches, our work specifically targets the underexplored domain of \textbf{3D semantic anomaly detection and recovery} to address high-level 3D structural inconsistencies and hallucinations.

\section{Limitation}
\begin{figure*}[t]
    \centering
    \includegraphics[width=\linewidth]{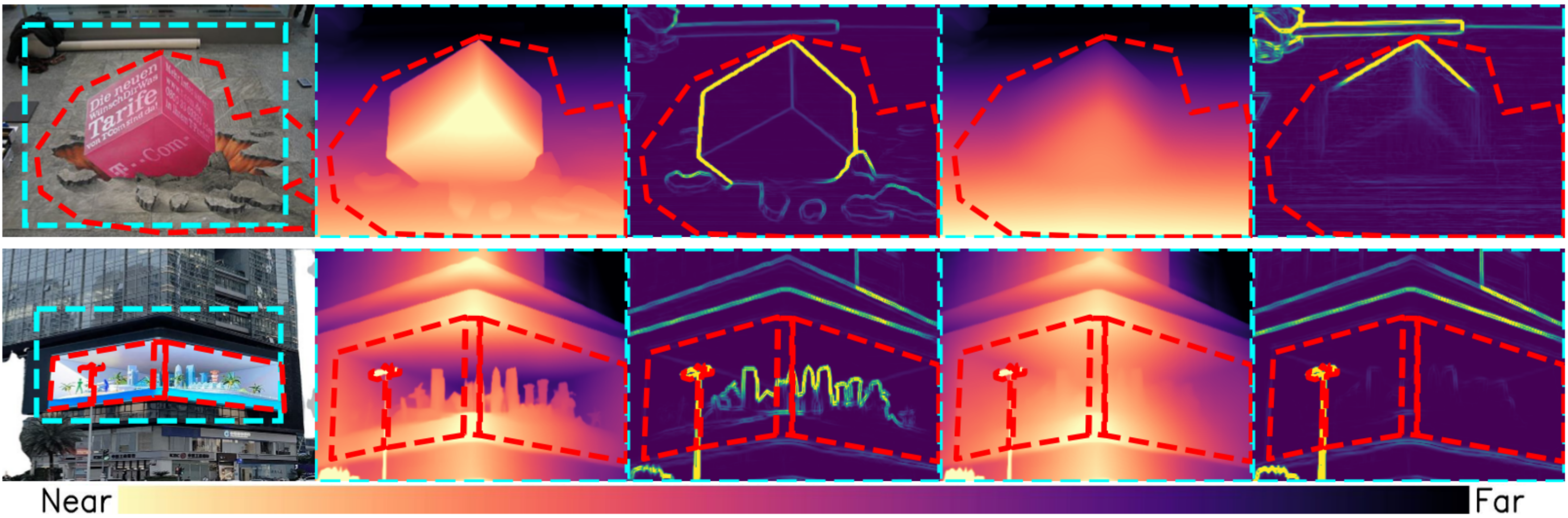}\vspace{-2mm}
    \caption{\textbf{Limitations and Failure Cases.} Examples where our model fails to fully suppress the 3D mirage. Left to right: Baseline Depth/Second-order, Ours Depth/Second-order. These failures often involve large protrusion illusions (e.g., cubes) or photorealistic scenes at long distances.}
    \label{fig:wefail}
\end{figure*}

There are some cases in which our model fails to recover expected depth (Fig.~\ref{fig:wefail}). The first case is likely because of the lack of Protrusion illusion in training data, especially one as prominent as the cube shown. Second case is harder to tackle as the scene looks photorealistic and is at long distance, and while our model successfully recover depth of multi-ROI/planes illusion before, this would require improved approach in future work.

\begin{figure*}[t]
    \centering
    \includegraphics[width=0.9\linewidth]{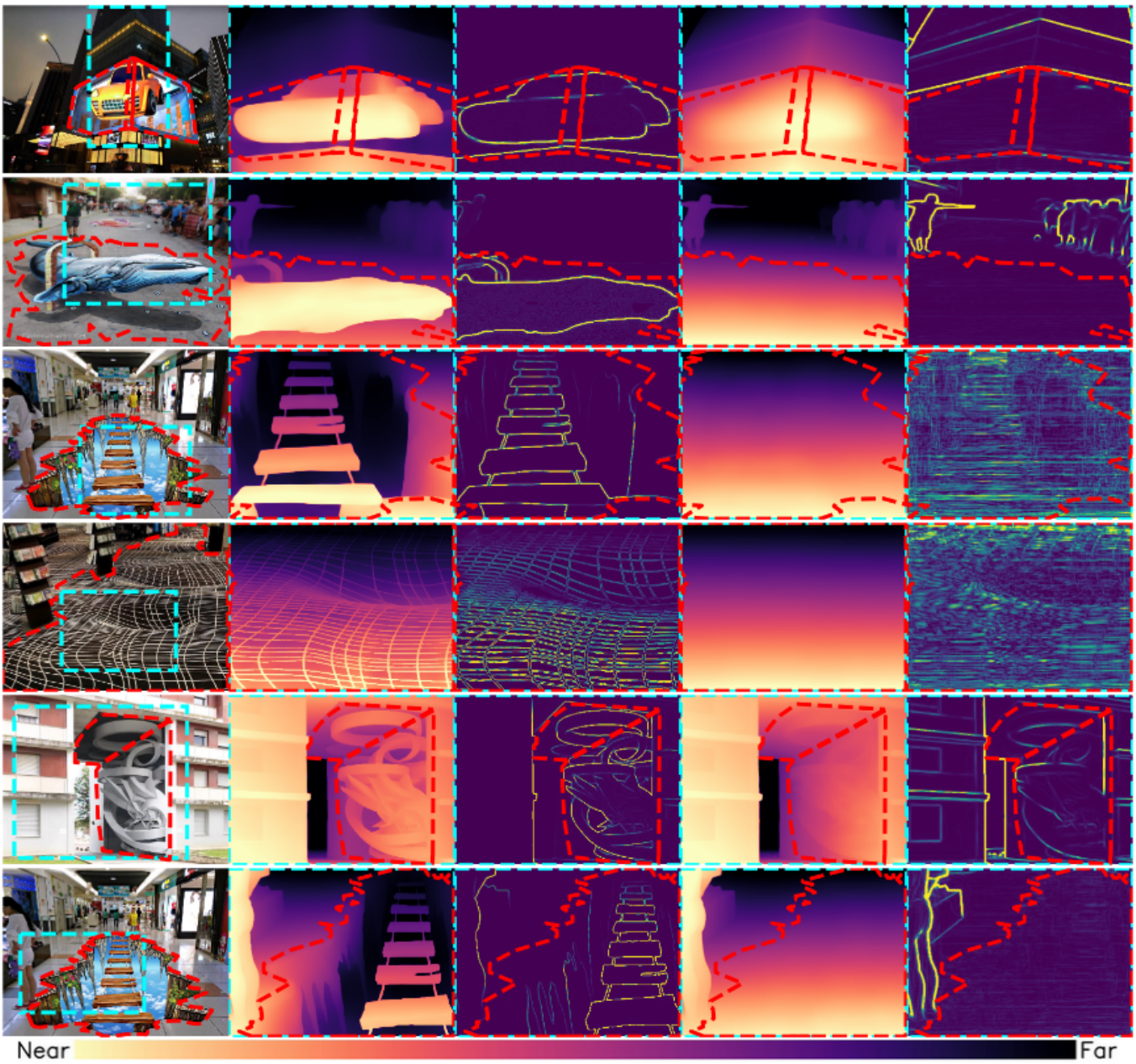}\vspace{-2mm}
    \caption{\textbf{Qualitative Results: Relative Models.} From left to right: Original RGB, depth-output and corresponding 2nd-order response of DA variants, then Ours. Rows 1-3 compare ours to DA(v1) Small-Base-Large (S-B-L), and rows 4-6 compare ours to DAv2 S-B-L. Our model successfully mitigates the top hallucination cases of the Depth Anything series under reduced context.}
    \label{fig:vs_exemplar_rel}
\end{figure*}

\begin{figure*}[t]
    \centering
    \includegraphics[width=\linewidth]{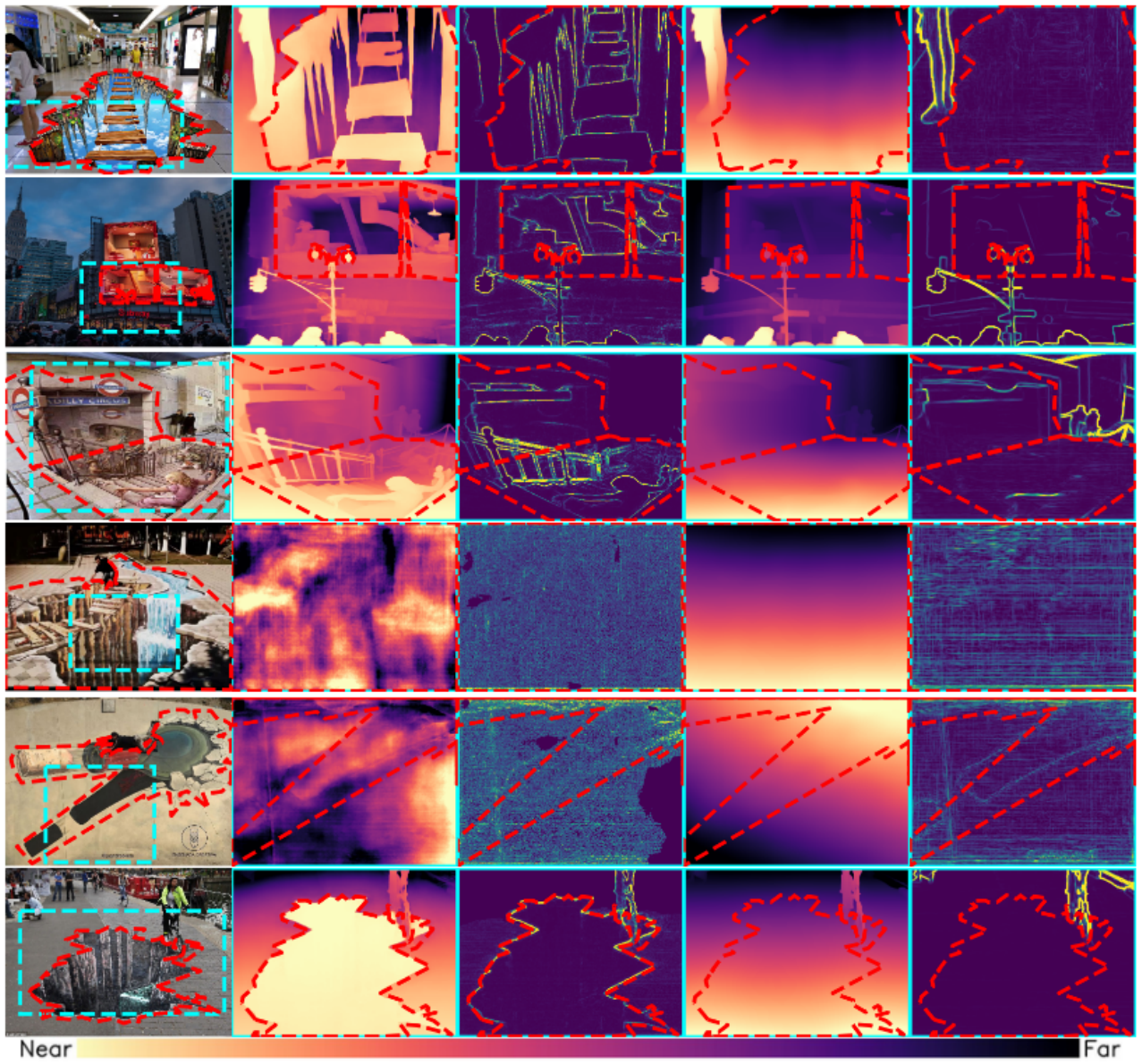}\vspace{-2mm}
    \caption{\textbf{Qualitative Results: Metric Models.}  From left to right: Original RGB, depth-output and corresponding 2nd-order response of DAv2 metric variants, then Ours. Rows 1-3 compare ours to DAv2-Indoor Small-Base-Large (S-B-L), and rows 4-6 compare ours to DAv2-Outdoor S-B-L. Our model mitigates hallucinations in DAv2 Indoor and Outdoor variants.}
    \label{fig:vs_exemplar_met}
\end{figure*}

\begin{figure*}[t]
    \centering
    \includegraphics[width=\linewidth]{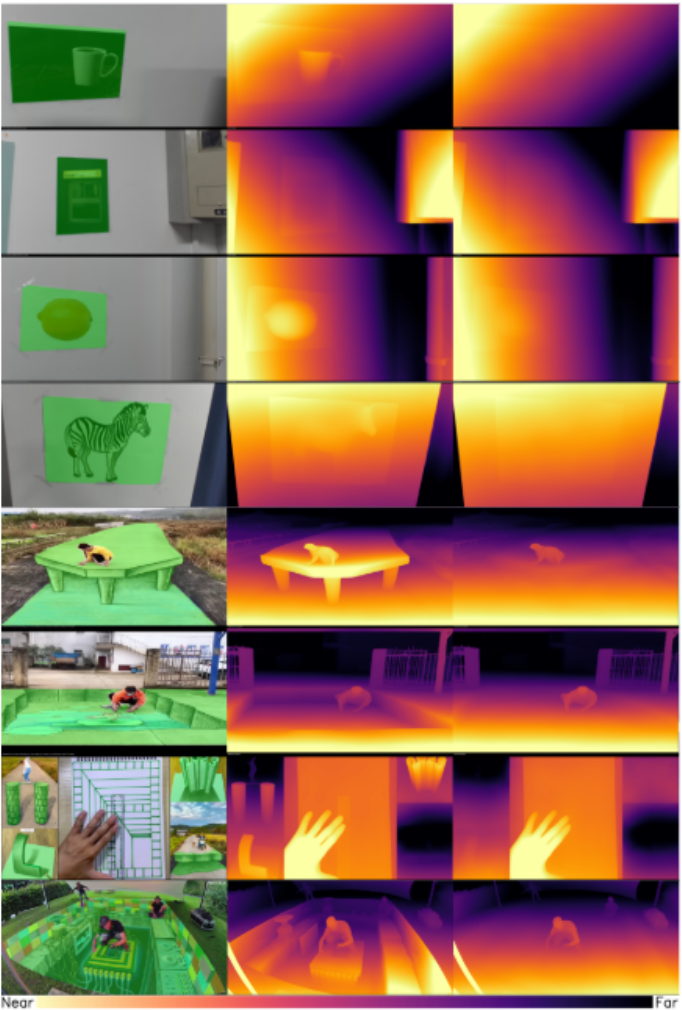}\vspace{-2mm}
    \caption{\textbf{Qualitative Results: Our model performance on 3D Visual Illusion dataset (Yao et al. NeurIPS’25) with no fine-tuning (zero-shot).}  From left to right: Original RGB, depth-output of baseline (Depth-Anything-V2 Large), then depth-output of Ours. Rows 1-4 come from their test set, and rows 5-8 come from their train set. Our model mitigates hallucinations compared to baseline.}
    \label{fig:ourson3dvi}
\end{figure*}

\noindent\textbf{Additional qualitative results.} We provide further side-by-side comparisons against baselines in Fig.~\ref{fig:vissup1}, Fig.~\ref{fig:vissup2}, and Fig.~\ref{fig:obworkin}, and extended results of our adapted model on relative, metric, and 3D-Visual-Illusion settings in Fig.~\ref{fig:vs_exemplar_rel}, Fig.~\ref{fig:vs_exemplar_met}, and Fig.~\ref{fig:ourson3dvi}.

\section{Cross-Architecture Transfer}
\label{sec:crossarch}
The 3D Mirage is not specific to the DAv2 encoder, and neither is its mitigation. We transfer Grounded Self-Distillation to two further backbones, a second transformer-based MDE (ZoeDepth) and a diffusion-based MDE (Marigold~\cite{Ke_2024_CVPR}), leaving the dual-view pipeline and the composite $\mathcal{L}_{\text{HKR}}+\mathcal{L}_{\text{NKP}}$ objective unchanged.

\vspace{0.4em}
\noindent\textbf{Second transformer backbone (ZoeDepth).} We inject LoRA adapters into ZoeDepth's patch-embedding projection and the encoder's attention-output and MLP linear layers. The adapted model improves consistently over its baseline across every metric (Table~\ref{tab:zoedepth_transfer}): DCS drops from \textbf{589.27 to 335.12} and $\delta_1$ rises from \textbf{76.16 to 84.76}. The scores remain weaker than our DAv2-L result, yet they confirm that the objective is not confined to a DINOv2-based encoder and yields directional robustness gains on a distinct transformer MDE.


\vspace{0.4em}
\noindent\textbf{Diffusion backbone (Marigold).} The 3D Mirage afflicts diffusion-based depth models as well (Fig.~\ref{fig:sota_hallucination}). Marigold builds on a frozen VAE encoder and decoder and a Stable-Diffusion-v2 U-Net under a single-step LCM scheduler; we inject LoRA adapters ($r{=}16$, $\alpha{=}32$) into every Linear and Conv2d module of the U-Net, training \textbf{16.2M} parameters ($\approx$1.87\% of the backbone) for a single epoch. As shown in Table~\ref{tab:marigold}, the adaptation transfers surgically: DCS falls from \textbf{2009 to 1127} and CCS from $\mathbf{77.5}$ to $\mathbf{46.9}$ ($\times10^{-4}$), a \textbf{44\%} and \textbf{40\%} reduction in hallucinated structure and contextual instability, while DA-2K and DIW preservation stay within one point of the Marigold baseline and zero-shot 3DVI improves ($\delta_1$ $83.73\rightarrow84.88$). We treat this as a feasibility check rather than a full diffusion-model study; it indicates that the objective transfers to an architecturally distinct backbone without modification.

\vspace{0.4em}
\noindent\textbf{Loss-weight sensitivity.} The result does not hinge on a tuned weighting. Beyond the default schedule (Sec.~\ref{sec:experiments}), Table~\ref{tab:marigold} reports a uniform setting ($\alpha_i{=}1$) and a mixed setting ($\alpha_i\in\{0,\tfrac{1}{2},1\}$) on DAv2-L; both hold DCS at $50.33$ and $49.26$ and CCS at $1.38$ and $1.51\times10^{-4}$, comparable to the default (DCS $58.64$, CCS $1.91\times10^{-4}$) with general-depth preservation unchanged.

\begin{table*}[t]
\centering
\caption{\textbf{Cross-architecture transfer and loss-weight sensitivity.} Grounded Self-Distillation on the Marigold diffusion backbone (bottom) cuts DCS and CCS by $44\%$ and $40\%$ while preserving general depth; uniform and mixed loss-weight settings on DAv2-L (top) match the default schedule, confirming robustness to $\{\alpha_i\}$. Lower is better except BG~$R^2$, DA-2K, DIW, and $\delta_1$. CCS is reported $\times10^{-4}$; NYU and KITTI are RMSE.}
\label{tab:marigold}
\setlength{\tabcolsep}{4pt}
\resizebox{\textwidth}{!}{%
\begin{tabular}{l|ccc|cccc|ccccc}
\toprule
& \multicolumn{3}{c|}{\textbf{3D-Mirage}} & \multicolumn{4}{c|}{\textbf{Standard Benchmarks}} & \multicolumn{5}{c}{\textbf{3DVI}} \\
\cmidrule(lr){2-4}\cmidrule(lr){5-8}\cmidrule(lr){9-13}
\textbf{Method} & DCS$\downarrow$ & CCS$\downarrow$ & BG $R^2\uparrow$ & NYU$\downarrow$ & KITTI$\downarrow$ & DA-2K$\uparrow$ & DIW$\uparrow$ & EPE$\downarrow$ & bad2$\downarrow$ & AbsRel$\downarrow$ & RMSE$\downarrow$ & $\delta_1\uparrow$ \\
\midrule
Ours (DAv2-Uniform) & 50.33 & 1.38 & 97.34 & 0.535 & 6.87 & 96.32 & 88.52 & 1.89 & 29.66 & 0.03 & 0.07 & 99.37 \\
Ours (DAv2-Mixed)   & 49.26 & 1.51 & 97.50 & 0.540 & 6.89 & 95.16 & 88.31 & 1.90 & 28.82 & 0.03 & 0.07 & 98.82 \\
\midrule
Marigold Baseline   & 2.009e3 & 77.5 & -- & 0.578 & 4.32 & 85.35 & 85.36 & 8.37 & 69.04 & 0.28 & 0.25 & 83.73 \\
Ours (Marigold)     & \textbf{1.127e3} & \textbf{46.9} & 94.06 & 0.636 & 4.52 & 84.77 & 85.00 & 8.02 & 70.55 & 0.23 & 0.22 & 84.88 \\
\bottomrule
\end{tabular}}
\end{table*}


\section{Annotation, Adaptation Cost, and Safety}

\noindent\textbf{Annotation protocol.} Every illusion is annotated with precise polygonal ROI masks; where a real object lies inside an illusion, a \emph{nested} ROI marks and excludes it, and a single scene may carry multiple illusions across distinct surfaces. All \textbf{1,872} full--crop instances were annotated and human-verified. During training, each ROI polygon's ring neighborhood supplies a small set of local plane hypotheses (the plane-mixture fit, Sec.~\ref{subsec:vis-rgs}), so re-editing is anchored to local geometry rather than a single global plane. The control study in Sec.~\ref{subsec:dbg-metrics} further shows DCS and CCS are insensitive to mild pixel and crop perturbations, so the metrics do not reward annotation jitter.

\noindent\textbf{Adaptation cost.} Grounded Self-Distillation is parameter-efficient: it trains only \textbf{4M} LoRA parameters ($\approx$0.7\% of the DAv2-L backbone) for a single epoch on one A100. The frozen teacher is used only during training; at inference the LoRA update folds into the backbone weights, so the adapted model keeps the \emph{same} latency and memory footprint as the unmodified baseline and adds no runtime overhead at deployment.

\noindent\textbf{Preserving genuine structure.} Because re-editing is gated to the annotated illusion ROIs and constrained by the $\mathcal{L}_{\text{NKP}}$ preservation loss, GSD suppresses spurious curvature without flattening real geometry outside the ROI, and the nested exclusions protect real objects embedded in an illusion. The \emph{No Knowledge Preservation} ablation (Sec.~\ref{sec:ablation_setup}) makes this explicit (removing $\mathcal{L}_{\text{NKP}}$ over-flattens and degrades general depth), confirming that preservation is what keeps genuine small structures intact. Validating this directly on real road hazards such as potholes and curbs is left to future work.

\bibliographystyle{splncs04}
\bibliography{main}
\end{document}